\documentclass{ecai} 

\usepackage{latexsym}
\usepackage{amssymb}
\usepackage{amsmath}
\usepackage{amsthm}
\usepackage{booktabs}
\usepackage{enumitem}
\usepackage{graphicx}
\usepackage{color}

\usepackage{subfigure}
\usepackage{epsfig}
\usepackage{makecell}
\usepackage{nicefrac}       
\usepackage{microtype}      
\usepackage{cases}
\usepackage{url}  
\usepackage{epstopdf}

\newtheorem{defn}{Definition}

\newcommand{\BibTeX}{B\kern-.05em{\sc i\kern-.025em b}\kern-.08em\TeX}

\begin{document}
	
	
	\begin{frontmatter}
		
		
		\paperid{M2002} 
		
		
		\title{Model Provenance via Model DNA}
		
		
		\author{\fnms{Xin}~\snm{Mu}}
		\author{\fnms{Yu}~\snm{Wang}}
		\author{\fnms{Yehong}~\snm{Zhang}}
		\author{\fnms{Jiaqi}~\snm{Zhang}}
		\author{\fnms{Hui}~\snm{Wang}}
		\author{\fnms{Yang}~\snm{Xiang}}
		\author{\fnms{Yue}~\snm{Yu}} 
		
		\address{Pengcheng Laboratory, Shenzhen, China}
		

		\begin{abstract}
			Understanding the life cycle of the machine learning (ML) model is an intriguing area of research (e.g., understanding where the model comes from, how it is trained, and how it is used).
			Our focus is on a novel problem within this domain, namely Model Provenance (MP). MP concerns the relationship between a target model and its pre-training model and aims to determine whether a source model serves as the provenance for a target model.
			In this paper, we formulate this new challenge as a learning problem, supplementing our exploration with empirical discussions on its connections to existing works.
			Following that, we introduce ``Model DNA'', an interesting concept encoding the model's training data and input-output information to create a compact machine-learning model representation. 
			Capitalizing on this model DNA, we establish an efficient framework consisting of three key components: DNA generation, DNA similarity loss, and a provenance classifier, aimed at identifying model provenance.
			We conduct evaluations on both computer vision and natural language processing tasks using various models, datasets, and scenarios to demonstrate the effectiveness of our approach. 
		\end{abstract}
		
	\end{frontmatter}
	

	\section{Introduction}
	
	In recent years, machine learning has become a ubiquitous presence in various fields, revolutionizing industries ranging from healthcare to finance \cite{DBLP:journals/cbm/ShehabASASAG22,dixon2020machine}. With the release of OpenAI's GPT-4, the capabilities of machine learning are poised to reach even greater heights\footnote{\url{https://openai.com/research/gpt-4}}. 
	However, as the value of data as an emerging asset class becomes increasingly recognized, the issue of understanding the entire life cycle of the model, e.g., understanding where the model comes from, how it is trained, and how it is used, has become a pressing concern.
	
	In this paper, we explore a novel and important problem within the research direction of understanding the machine learning model life cycle, namely \textbf{Model Provenance} (MP).
	This problem aims to investigate the relationship between two models, such as understanding whether a target model is derived from a source model. 
	To illustrate this problem, consider a real-world business scenario in which an AI company seeks to protect the intellectual property of its machine learning model trained on private data using a significant amount of computing power. The company wishes to ensure that the model is only used by authorized users. However, in practice, authorized users may share the model with unauthorized users without the company's permission, or the model may be stolen and used by an unauthorized user. The unauthorized user may then use the stolen model to develop their own products based on techniques such as continual learning \cite{DBLP:journals/nn/ParisiKPKW19}.
	This situation presents a significant challenge: how can an AI company identify whether a source model, owned by an authorized user, is the provenance of a target model developed by an unauthorized user?
	
	A relevant area of study involves investigating the influence of pre-training models in continual learning or lifelong learning. For instance, recent work by \cite{DBLP:journals/corr/abs-2112-09153} has shown that generic pre-training can implicitly counteract the negative effects of catastrophic forgetting when learning multiple tasks sequentially, particularly when compared to models initialized randomly. Additionally, research conducted by \cite{DBLP:conf/iclr/TonevaSCTBG19} highlights the occurrence of catastrophic forgetting in the context of continual learning scenarios. Furthermore, certain works have explored connections between the outputs of different models \cite{DBLP:conf/nips/CsiszarikKMPV21,DBLP:conf/icml/Kornblith0LH19}.
	While previous research has explored the relationship across various tasks to understand the phenomenon of catastrophic forgetting or the outputs of different models, there is currently no established method for identifying the relationship between different models across diverse tasks, to the best of our knowledge. 
	
	In this paper, we address the problem of model provenance and begin by formalizing this problem and conducting an empirical study to evaluate the performance of the target model on the source model's training data (Section~\ref{sec:def}), 
	whose results inspire us to ask whether we can create a description to explain the relationship between the source and target models. 
	Then, we introduce a novel concept of model DNA, which represents the unique characteristics of a machine learning model, and propose a framework for model DNA generation and model provenance identification. The effectiveness of our approach is demonstrated through both Computer Vision (CV) and Natural Language Processing (NLP) tasks. 
	
	The contributions of the paper are summarized as follows: 
	\begin{itemize} 
		\item This paper investigates a critical aspect of understanding the machine learning models' life cycle - identifying the homologous relations between a source model and its subsequent target versions. We first formulate this challenge as a learning problem and provide empirical discussions on its relation to existing works.
		
		\item  We propose a novel machine learning model representation, called Model DNA, which combines data-driven and model-driven approaches to encode the training data and input-output information as a compact representation of the model (Section~\ref{sec:dna}). In DNA space, we can easily measure the similarity between models and track their usage and evolution over time. Based on this idea, we introduce the Model Provenance framework, which includes Model DNA generation and provenance identification, providing a practical solution for identifying relationships between models and better understanding the provenance of machine learning models. 
		\item  We perform experiments on various commonly used CV and NLP benchmark datasets along with DNN structure models to assess the effectiveness of our proposed framework. To enhance the understanding, we present comprehensive visualizations of the distribution of Model DNA in the DNA space (Section~\ref{sec:exp}).
	\end{itemize}
	
	\section{Related work}
	\textbf{Lifelong learning.} 
	Lifelong machine learning is a learning paradigm that continuously accumulates past knowledge and adapts it to future learning and problem-solving \cite{DBLP:journals/pami/LangeAMPJLST22,DBLP:series/synthesis/2018Chen,DBLP:journals/nn/ParisiKPKW19}. This involves performing a sequence of $N$ learning tasks, $\mathcal{T}_1, \mathcal{T}_2,\ldots, \mathcal{T}_N $, each with its corresponding dataset
	$D_1, D_2,\ldots, D_N$
	of different types and from different domains. When faced with a new task $\mathcal{T}_{N+1}$ (called the current task) with its data $D_{N+1}$, the learner can use past knowledge to aid in learning.
	One of the major challenges in lifelong learning is catastrophic forgetting, where the model forgets previously learned information when learning new tasks. To mitigate this phenomenon, approaches can be categorized into three groups: (1) regularization-based approaches \cite{kirkpatrick2017overcoming,DBLP:conf/icml/ZenkePG17}, (2) memory-based approaches \cite{DBLP:conf/nips/Lopez-PazR17,DBLP:conf/iclr/ChaudhryRRE19,DBLP:conf/emnlp/WangMPC20}, and (3) network expansion-based approaches \cite{DBLP:journals/corr/RusuRDSKKPH16,DBLP:conf/cvpr/AljundiCT17,DBLP:journals/neco/SodhaniCB20}.
	This paper aims to distinguish the relationship between the model from the previous task and the current task, which is an under-explored direction in existing lifelong learning works.
	
	\textbf{Representation Learning.} Representation learning refers to the process of learning a parametric mapping from the raw input data domain to a feature vector, in the hope of capturing and extracting more abstract and useful concepts that can improve performance on a range of downstream tasks. The aim of learning representations of the data is to make it easier to extract useful information when building classifiers or other predictors. 
	Representation learning has played a tremendous role in different frameworks \cite{DBLP:journals/pami/BengioCV13}. Recently, Contrastive Representation Learning (CRL) is widely used in NLP and CV \cite{DBLP:journals/access/Le-KhacHS20}. It can be considered as self-supervised learning by comparing among different samples \cite{DBLP:conf/icml/ChenK0H20,DBLP:journals/corr/abs-1807-03748}.
	In this paper, we explore a way of machine learning model representation. Our approach involves a data-driven and model-driven representation learning framework that constructs a model representation in a latent space. Additionally, there are some works to investigate the relationship of Neural Network representations \cite{DBLP:conf/nips/CsiszarikKMPV21,DBLP:conf/icml/Kornblith0LH19}. This work is novel since it is the first time, as far as we know, that such an approach is proposed for representing the ML model.
	
	\textbf{Data Provenance.} 
	The area of data provenance is also relevant and falls under the provenance research family. In general, the problem is defined by auditing if a certain piece of data has been used to train a machine learning model \cite{DBLP:conf/kdd/SongS19,DBLP:journals/corr/abs-2209-01538,DBLP:conf/iclr/MainiYP21}. It is also called \textit{membership inference attacks} \cite{DBLP:conf/sp/ShokriSSS17,DBLP:journals/corr/abs-2103-07853}.
	A data provenance technique (i.e., shadow training) has been widely studied, which can successfully audit deep learning-based models \cite{DBLP:conf/sp/ShokriSSS17,DBLP:conf/kdd/SongS19}. The main idea is to use multiple shadow models to imitate the behavior of the target model. As the training data for the shadow models are known, the target model can be trained using the labeled outputs of the shadow models. However, this is a data-level provenance problem, whereas our focus is on the model-level provenance problem.

	\section{Model Provenance (MP)}\label{sec:def}
	
	\subsection{Definition} 
	
	Consider a dataset $D_s = \{(x_i,y_i)\}_{i=1}^{{|D_s|}}$ where each $x_i \in \mathbb{R}^d$ is a data instance and $y_i \in Y =\{1, 2,\ldots,c\}$ is its associated class label, a machine learning model $M_s$ is learned from $D_s$ by using an algorithm architecture $\mathcal{A}_s$.
	Let $D_t = \{(x_j,y_j)\}_{j=1}^{{|D_t|}}$ be a dataset used to learn a machine learning model $M_t$. If $M_t$ is learned using the pre-training model $M_s$ with the same algorithm architecture $\mathcal{A}_s$, we refer to $M_s$ as the source model and $M_t$ as the homologous model of $M_s$.
	Similarly, let $\bar{M}_t$ be a machine learning model learned from the dataset $D_t = {(x_j,y_j)}_{j=1}^{{|D_t|}}$ but based on random initialization or pre-training without using $M_s$. 
	We refer $\bar{M}_t$ as non-homologous to $M_s$.
	
	The problem of \emph{Model Provenance (MP)} is to find a function $f$ such that for a given source model $M_s$ and any target model $M_t$, 
	\begin{equation} 
		f(M_s, M_t) = \begin{cases} 
			1,  & \mbox{if $M_t$ is the homologous model of $M_s$.} \\
			0, &  \mbox{otherwise}.
		\end{cases}
	\end{equation}
	When $f(M_s, M_t) = 1$, we say $M_s$ is the provenance of $M_t$. 
	
	\begin{figure*}[t]
		\centering
		\subfigure[]{\epsfig{file=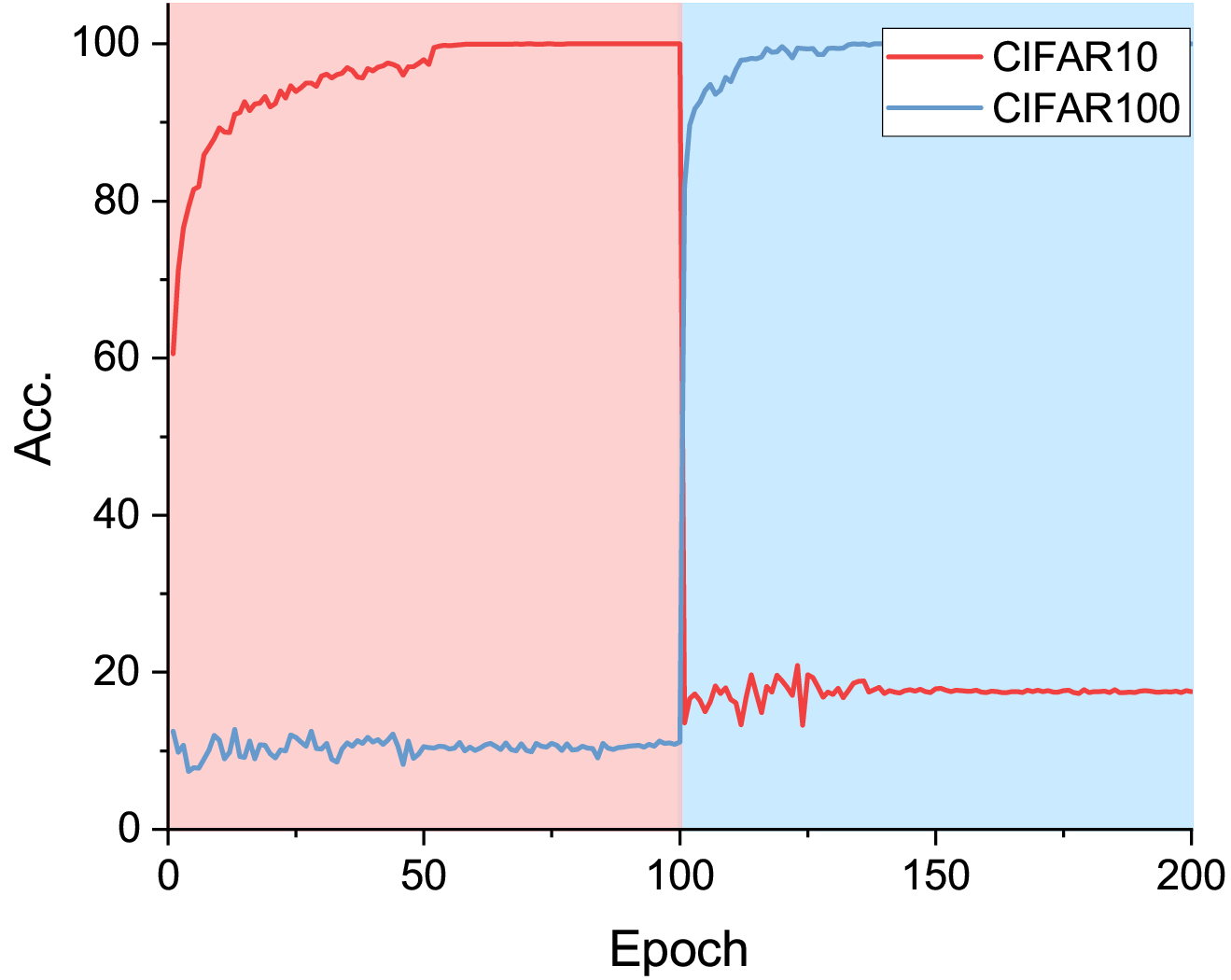,height=1.35in,width=1.35in}} 
		\subfigure[]{\epsfig{file=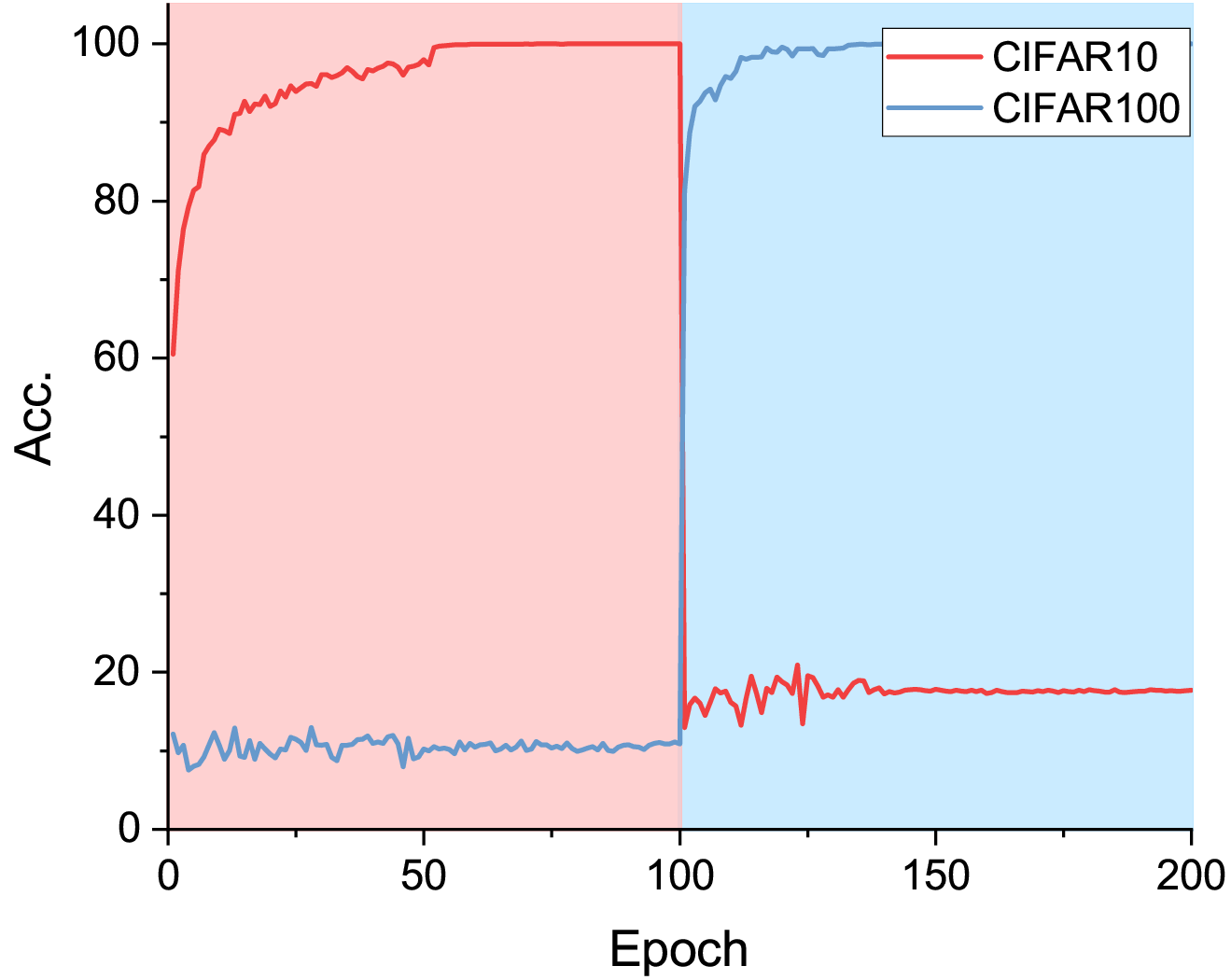,height=1.35in,width=1.35in}} 
		\subfigure[]{\epsfig{file=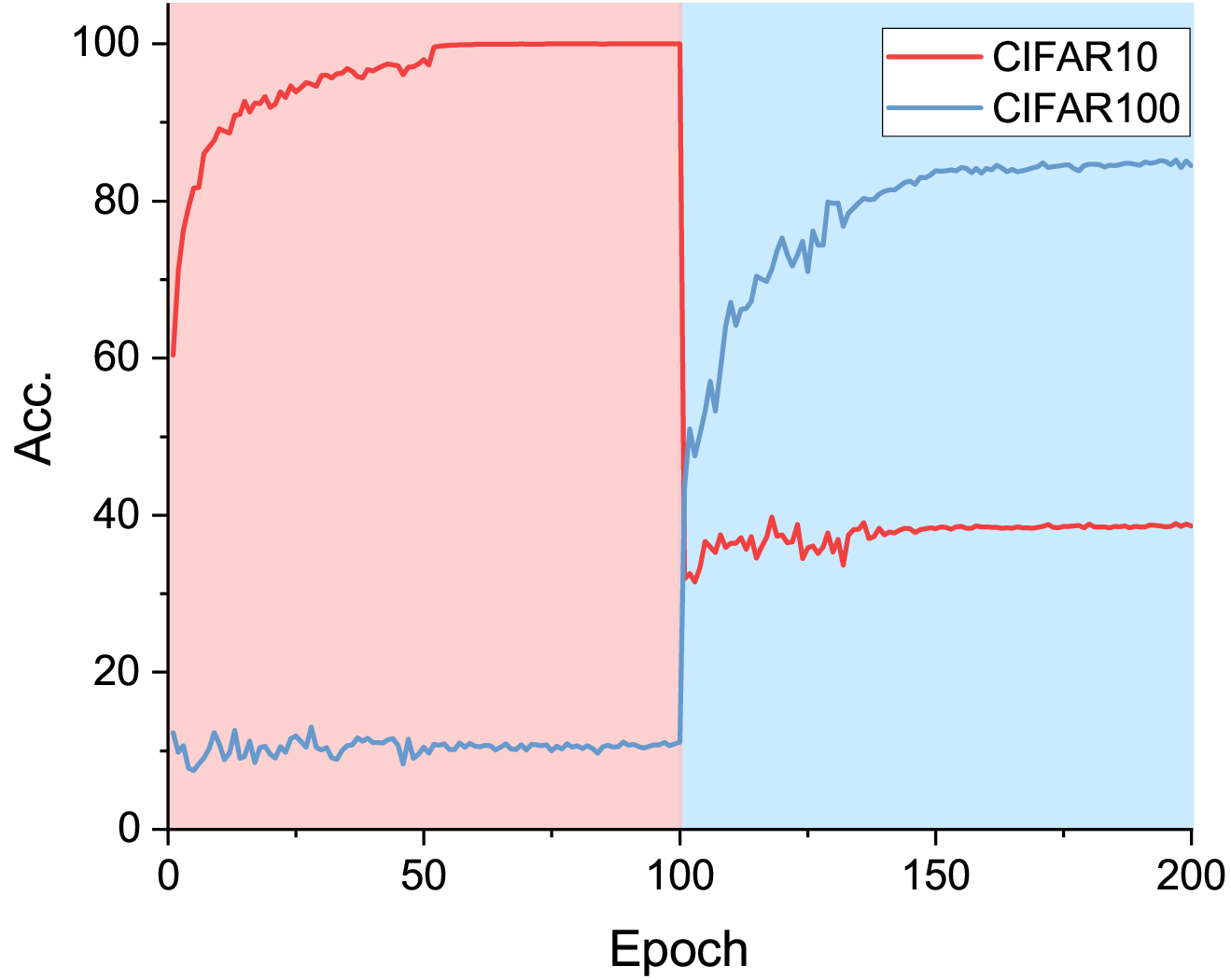,height=1.35in,width=1.35in}} 
		\subfigure[]{\epsfig{file=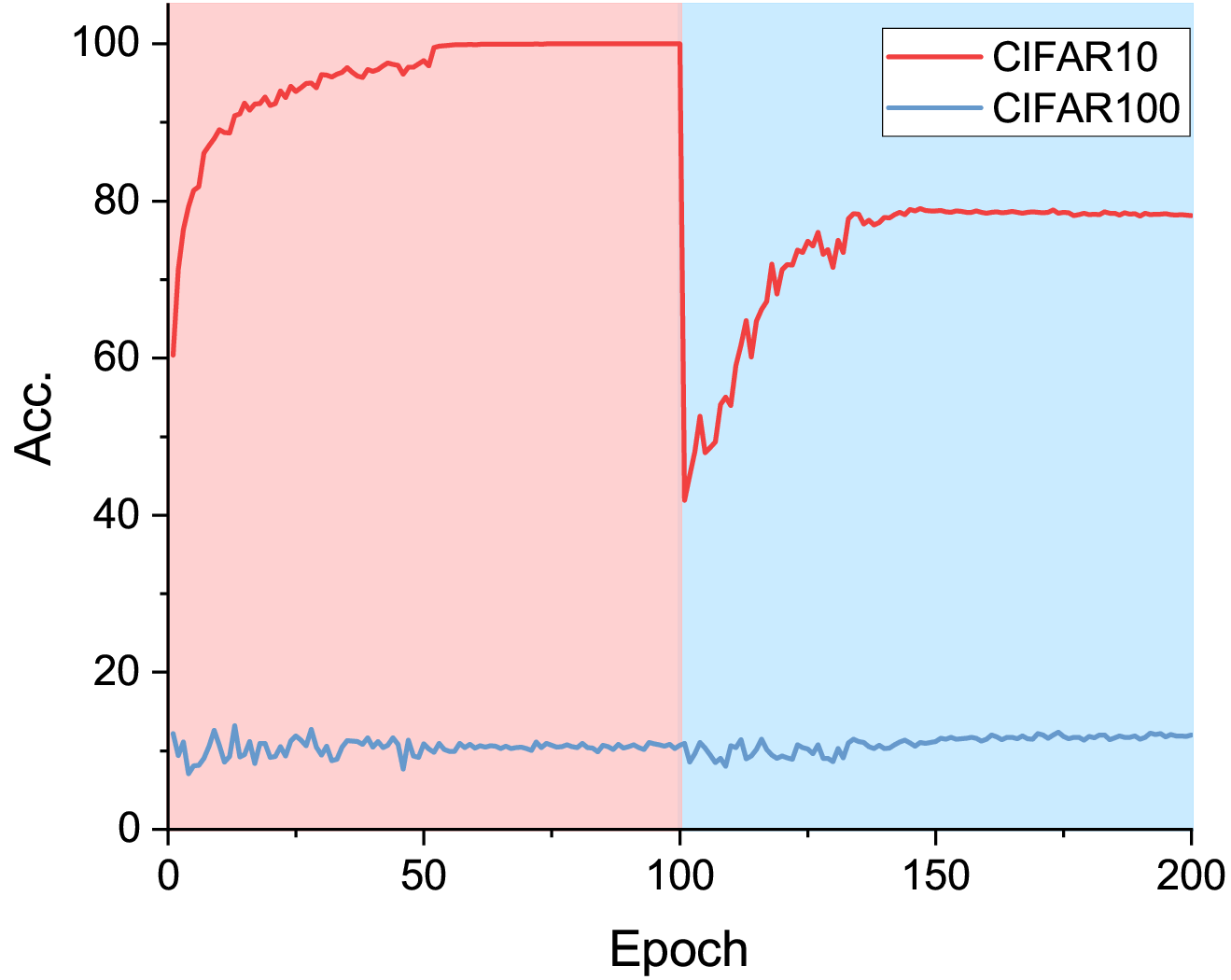,height=1.35in,width=1.35in}} 
		\subfigure[]{\epsfig{file=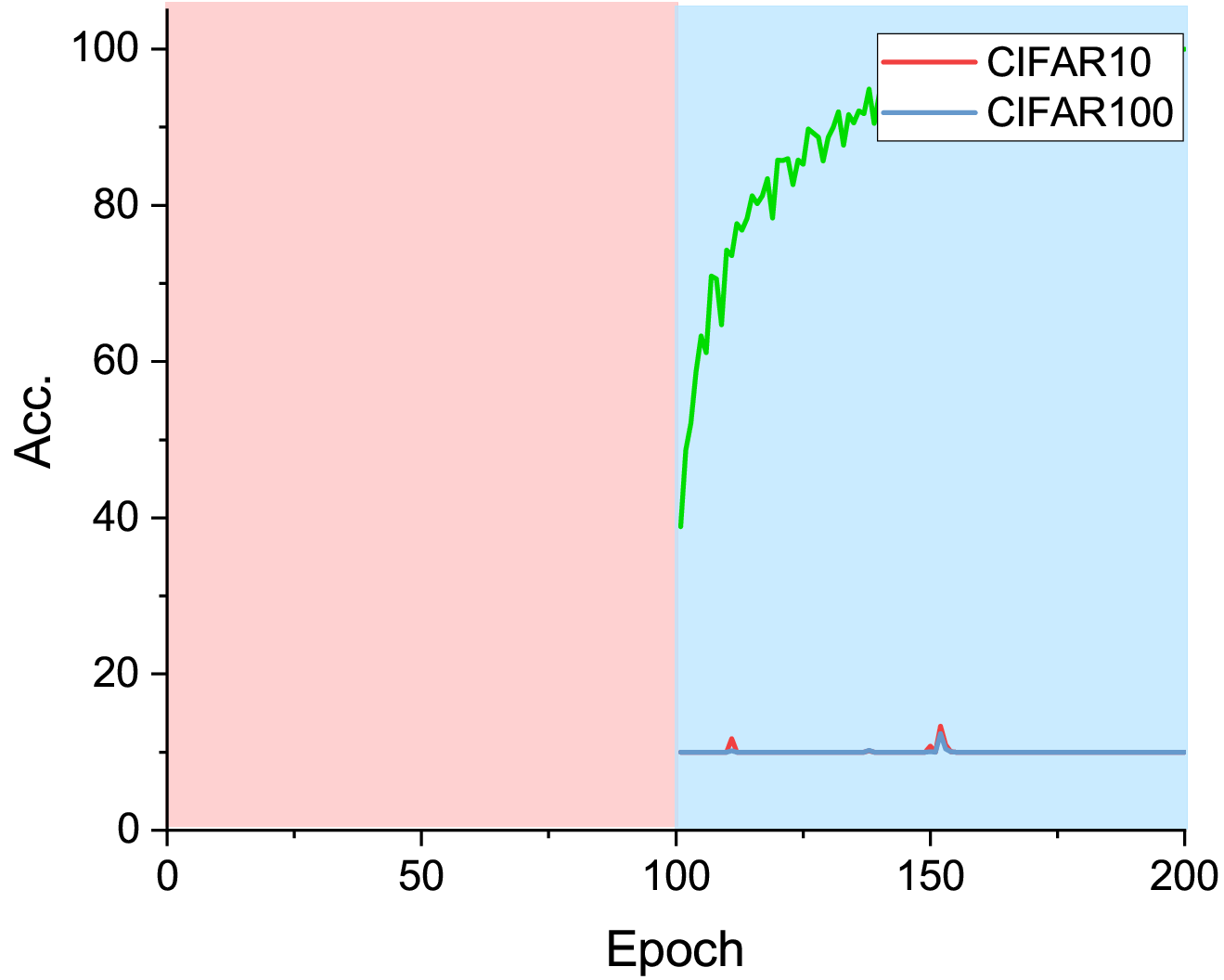,height=1.35in,width=1.35in}} 
		\caption{ResNet18. (a) No replace layer. (b) Replace target model's last layer with source model. (c) Replace last two layers. (d) Replace last three layers. (e) Replace target model (random initialization)'s last three layers with source model.}
		\label{fig:1}
	\end{figure*}
	
	\vspace{1mm}
	\textit{Remark}: 
	The problem of Model Provenance (MP) can have various variations based on the conditions applied to the target model $M_t$. 
	For example, there would be a lot of downstream models obtained by fine-tuning, distillation, pruning, etc. 
	In this paper, our primary focus is on the MP scenario within the fine-tuning setting, specifically when $M_t$ shares the same structure as $M_s$, and its training solely depends on the pre-training model $M_s$. Our exploration of MP is initially focused on this easier condition, and we intend to explore more complex scenarios in future research.
	
	\subsection{Discussion} 
	
	In the field of continual learning, there have been several studies focused on quantifying the relationship between homologous models, such as the analysis of catastrophic forgetting \cite{DBLP:conf/nips/MirzadehFPG20,DBLP:conf/iclr/KeskarMNST17,DBLP:journals/corr/abs-2112-09153}. For example, \cite{DBLP:conf/nips/MirzadehFPG20} provided a theoretical bound on forgetting in sequential learning. Let $L(w)$ be the loss on the training dataset with model parameter $w$, and $w_1$ and $w_2$ be the optimal or convergent parameters after training the source and homologous models sequentially. Specifically, they showed that
	\begin{equation}
		L_1(w_2)-L_1(w_1)\approx\frac{1}{2}\Delta w^{\top}\nabla^2 L_1 (w_1) \Delta w \leq \frac{1}{2} \lambda_1^{max} ||\Delta w||
		\label{them_1}
	\end{equation}
	where $L_1(w_1)$ and $L_1(w_2)$, respectively, represent the losses on source training dataset with parameters $w_1$ and $w_2$, $\Delta w = w_2 - w_1$, and $\lambda_1^{max}$ is the largest eigenvalue of the Hessian matrix $\nabla^2 L_1 (w_1)$. The eigenvalues of the Hessian matrix are indicative of the curvature of the loss function \cite{DBLP:conf/iclr/KeskarMNST17}, where smaller eigenvalues imply a flatter loss function.
	Thus, $\lambda_1^{max}$ is considered as a proxy for the flatness of the loss function (lower is flatter). Meanwhile, empirical studies conducted by \cite{DBLP:journals/corr/abs-2112-09153} and \cite{DBLP:conf/nips/MirzadehFPG20} show that initializing models with pre-trained weights results in a relatively flat task minima, and flatter models tend to have smaller $\Delta w$ than less flat ones.

	These existing works have motivated us to a straightforward solution for model provenance, i.e., using the difference of $\Delta w$ to distinguish between homologous and non-homologous models. We can expect that $\Delta w = w_2 - w_1$ would be smaller for homologous models compared to non-homologous models based on the above analysis.
	However, we found that this solution did not work well in practice. The main issue was that it is impossible to know how flat the target model is, which makes it difficult to use the difference of $\Delta w$ as a reliable indicator of model provenance.

	Furthermore, we conducted experiments using the standard continual learning setup on the image classification task~\cite{DBLP:conf/iclr/TonevaSCTBG19,DBLP:conf/nips/PaulGD21} to show the relationship between homologous models. Specifically, we examined two model architectures, ResNet18~\cite{DBLP:conf/cvpr/HeZRS16} and AlexNet~\cite{DBLP:conf/nips/KrizhevskySH12}, on the CIFAR10 and CIFAR100 datasets. A source model $M_s$ is first trained on CIFAR10. Then we randomly select ten classes from CIFAR100 to train target models $M_t$. Note that $M_t$ is learned by initializing its parameters using those of the source model $M_s$. In Figure \ref{fig:1}, we present the results of the experiments conducted on ResNet18, and the results for AlexNet can be found in \cite{mu2023model}. The background color indicates the different training phases. The red color represents the training of source model $M_s$ on the CIFAR10 dataset, while the blue color represents the training of target model $M_t$ on the CIFAR100 dataset. The red line corresponds to the accuracy of testing the model on the training data of CIFAR10, while the blue line represents the accuracy of testing the model on the training data of CIFAR100. In the blue background part, we train the model based on the parameters of the model obtained in the red background part.
	
	Figure \ref{fig:1} (a) demonstrates that the predictive accuracy of $M_t$ is significantly reduced when evaluated on the CIFAR10 dataset. The red section represents a model (called source model) trained on the source domain (CIFAR10), while the blue section represents this model (called target model) continuously trained on the target domain (CIFAR100). The red line indicates model evaluation on CIFAR10, and the blue line represents model evaluation on CIFAR100.
	This observation suggests that the training data of the source model may have been severely forgotten in the homologous model (i.e., large $||\Delta w||$). 
	Figures \ref{fig:1} (b)-(d) provide further insight into the relationship between the source model $M_s$ and the homologous model $M_t$. They show the predictive accuracy of $M_t$ on the CIFAR10 dataset when different numbers of the last layers of $M_t$ are replaced with corresponding layers in $M_s$.
	In Figure \ref{fig:1} (b), within the blue section, we replace the model's last layer with the model from the red section. Consequently, the model exhibits reduced performance when evaluated on CIFAR10 (red line). Similarly, in Figures \ref{fig:1}~(c)~and~(d), we extended this by replacing the model's last two/three layers with the ones from the red section, respectively.
	We can observe that as we replace more layers in $M_t$ with those from $M_s$, the predictive accuracy of $M_t$ on the CIFAR10 dataset increases.
	In Figure \ref{fig:1} (e), we consider a scenario where in the blue section, the (target) model is not initialized by the (source) model from the red section. Instead, the (target) model is trained with random initialization. In this setting, we replace the model's last three layers with the corresponding layers from the (source) model in the red section (as depicted in Figure \ref{fig:1} (a)). Figure \ref{fig:1} (e) shows that even with the last layers replaced, the predictive accuracy of $\bar{M}_t$ is still poor. The green line in Figure \ref{fig:1} (e) represents the training accuracy of $\bar{M}_t$.
	Figures \ref{fig:1} (b) and (e) illustrate the presence of a relationship between the source model and its homologous model. However, non-homologous models, do not exhibit any discernible relationship even by replacing the layers between them.
	
	Overall, our discussion has demonstrated the existence of a relationship between the source model and its homologous model, as shown by the improvement in predictive accuracy when replacing the last layers of the target model with those of the source model. 
	These results led us to pose the question: ``\emph{Can we establish a relationship between a source model and its homologous or non-homologous model based on the training data of the source model?}'' and motivated us to introduce the ``Model DNA'' and Model Provenance framework. 
	
	\section{The proposed framework}\label{sec:framework}
	
	\subsection{Model DNA}\label{sec:dna}
	
	In the field of biology, deoxyribonucleic acid (DNA) is known as the molecule responsible for carrying genetic information essential for the growth and operation of organisms \cite{ridley2000genome}. In the context of Machine Learning (ML), we introduce the concept of model DNA as a form of representation for an ML model. Drawing parallels to prior research in representation learning \cite{DBLP:journals/pami/BengioCV13}, this model representation aids in identifying differences among various model iterations across diverse tasks. Furthermore, it facilitates comparisons and assessments of similarity between different models.
	
	In the domain of machine learning, a typical process involves training an ML model using a dataset $D$ and an algorithm architecture $\mathcal{A}_s$. With this understanding, our approach to DNA generation factors in both the impact of the dataset $D$ and the model's input-output relationship. Thus, we present the model DNA definition:
	\begin{defn}{\textbf{(Model DNA)}:} Let $D$ be a set of $N$ training data of a machine learning model $M$. 
		We define the model DNA as a set of $N$ DNA fragment $\mathbb{O}=\{o_1,o_2,\ldots,o_N\}$, where each DNA fragment $o_i$ is corresponding to a training sample of the model. It can be generated as $\mathbb{O} \leftarrow g(D,M)$
		where $g(\cdot)$ is an approach for DNA generation. A DNA fragment $o_i$ is generated by $g(x_i,M)$ for $x_i \in D$.
		\label{define:DNA}
	\end{defn}

	Through the conceptualization of model DNA, we create a latent space of model representations encompassing the diverse DNA of various ML models. In this space, we can quantify the relationships between different ML models.
	Here, we assume that a DNA fragment $o_i$ is a vector by $o_i \leftarrow g(x_i, M)$, where $x_i \in D$. 
	It is important to note that we generally assume that the DNA of homologous models will be positioned closer, while the DNA of non-homologous models will be relatively distant from each other. This latent space leads to several key properties of model DNA, as described below.
	
	\begin{figure*}[t]
		\centering
		\includegraphics[width=15cm,height=6cm]{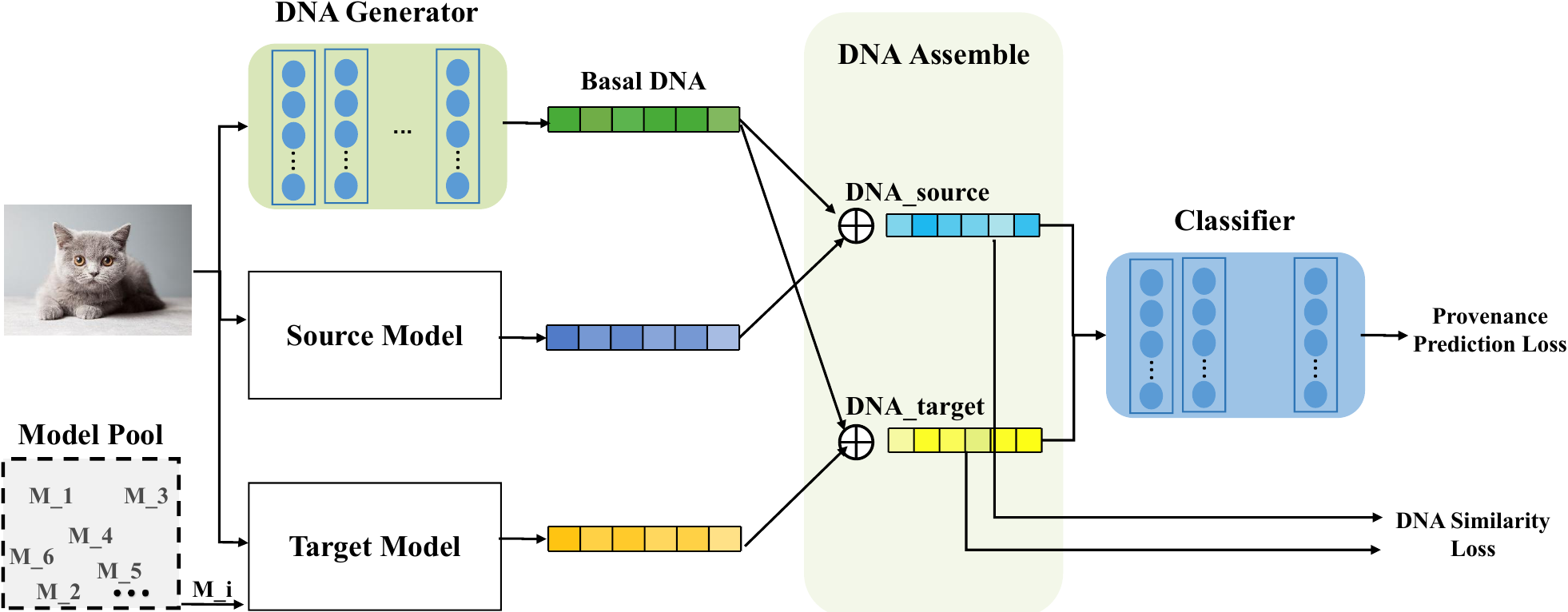}
		\caption{The MGMP framework.}
		\label{fig:my_model}
	\end{figure*}

	Let $\mathbb{O}_i$ be the model DNA of any model $M_i$ and $\mathbb{D}$ represent the distance function such that $\mathbb{D}(o^s_i,o^t_j)$ is the distance between DNA fragments of any two models $M_s$ and $M_t$. The following properties of DNA latent space are desired:
	
	\begin{itemize}[leftmargin=15pt,itemsep=5pt]
		
		\item Different ML models have different DNA: If $M_s \neq M_t$, $\mathbb{O}_s \neq \mathbb{O}_t$.

		\item The homologous models should have similar DNA fragments and vice versa: $\mathbb{D}(o^s_i,o^t_i) < \mathbb{D}(o^s_i,\bar{o}^t_i)$, where $o^s_i \leftarrow g(x_i,M_s)$, $o^t_i \leftarrow g(x_i,M_t)$ and $\bar{o}^t_i \leftarrow g(x_i,\bar{M}_t)$. $M_t$ is trained based on the pre-training model $M_s$, whereas $\bar{M}_t$ is trained without using $M_s$.
		
		\item The DNA fragments from the same ML model should be similar: $\forall i,j \ , i\neq j$, $\mathbb{D}(o_i,o_j)$ should be minimized for $o_i,o_j \in \mathbb{O}$.
		
	\end{itemize}
	
	In the following, we present our framework aimed at tackling the MP problem. This framework has been meticulously crafted to adhere to the aforementioned model DNA properties. Our approach takes into account not only the training data of the source model but also the model's input-output to produce a comprehensive model DNA representation. 
	
	\subsection{MGMP}\label{sec:framework1}
	
	We introduce a framework termed Model DNA Generation and Model Provenance identification (\textbf{MGMP}), which operates on the principles of deep representation learning. 
	The \textbf{MGMP} framework comprises three primary components: DNA generation, DNA similarity loss, and provenance classifier, as depicted in Figure \ref{fig:my_model}. The process commences with the source model's training data, which is fed into the DNA generator alongside the source and target models. This concatenation produces outputs from the DNA generator, source model, and target model, which are then amalgamated to generate the model DNA, exemplified by the integration of DNA generator outcomes and model predictions. 
	Subsequently, we formulate the model DNA for both the source and target models. To preserve information relevant to homologous and non-homologous models, a DNA similarity loss is incorporated. Finally, we exploit the model DNA to train a binary prediction network, allowing us to infer provenance outcomes. Each component of the \textbf{MGMP} is described in subsequent sections.

	\subsubsection{DNA generation}

	The process of DNA generation involves a combination of data-driven and model-driven techniques aimed at acquiring a representation for an ML model.
	Inspired by \cite{DBLP:journals/cacm/GoodfellowPMXWO20}, 
	we initiate the process by initializing a generator $g(\cdot)$ using a standard deep neural network architecture (it's important to emphasize that the generator could potentially be constructed using more intricate deep generative models). This generator encompasses multiple hidden layers, employing the rectified linear unit (ReLU) activation function.

	Given an input data point $x_i$, the generator produces a latent representation denoted as $z_i=g(x_i)$ (e.g., the output of the last layer). Here, $g(\cdot)$ signifies the learned DNA generation function realized through the deep network. The \textbf{MGMP} framework utilizes the training data $D_s$ of the source model as its own training data. Concretely, each training data point $x_i$ is individually fed into the DNA generator, source model, and target model,  as follows:
	\begin{equation}
		z_i \leftarrow g(x_i),  \ \ \ y^s_i \leftarrow M_s(x_i), \ \ \ y^t_i \leftarrow M_t(x_i)\ .
	\end{equation}

	\textbf{Model pool.} 
	The primary objective of the proposed framework is to establish a representation space, enabling the comparison of distinct models, encompassing both those that are homologous and those that are not. To realize this objective, we utilize a model pool that encompasses a variety of models, each demonstrating different relationships with the source model. In each training mini-batch, we select a pair of models $(M_t,\bar{M_t})$ to represent the target model (as shown in Figure \ref{fig:my_model}), where $M_t$ is learned based on the source model $M_s$, while $\bar{M_t}$ is learned based on random initialization.

	For each input $x_i$, the framework produces four distinct outputs. These include the foundational DNA representation $z_i$, the output $y^s_i$ from the source model, and the outputs $y^t_i$ and $\bar{y}^t_i$ from the homologous and non-homologous target models respectively.

	\textbf{DNA assemble.}
	After that, the outputs of the DNA generator and the source/target models are merged to create a model DNA fragment for each input $x_i$. The specific approach for combining these outputs can vary. In this study, we assume that the outputs are of the same dimension and are combined through addition. As a result, for each input $x_i$, we can generate three distinct types of DNA fragments:
	\begin{equation}
		o^s_i \leftarrow z_i + y^s_i,  \ \
		o^t_i \leftarrow z_i + y^t_i,   \ \bar{o}^t_i \leftarrow z_i + \bar{y}^t_i \ . 
	\end{equation}

	\subsubsection{DNA similarity loss}
	
\begin{figure}[t]
		\centering
		\includegraphics[scale=0.6]{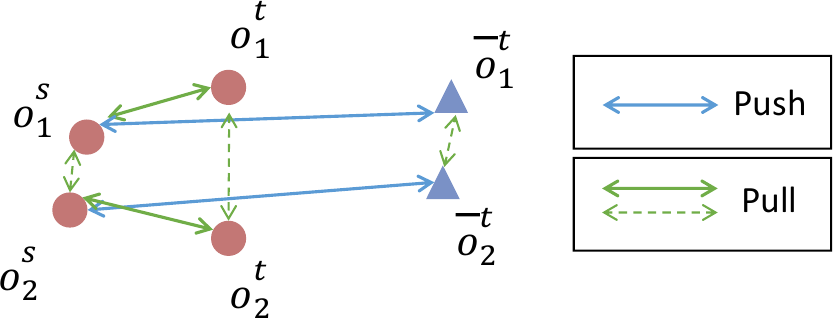}
		\caption{Similarity in DNA space.}
		\label{fig:SDC}
	\end{figure}

	The DNA similarity loss is designed with a specific purpose: to measure the similarity between the DNA of two models and to satisfy the properties of DNA latent space. 
	We achieve this by employing a metric that ensures that the similarity of $(o^s_i, o^t_i)$ in the latent space is not disclosed to $(o^s_i, \bar{o}^t_i)$, which is inspired by Contrastive Learning \cite{DBLP:conf/icml/ChenK0H20,DBLP:conf/cvpr/Yuan0K0WMKF21}. In more detail, our approach does not just preserve the relationships like $(o^s_i, o^t_i)$ and $(o^s_i, \bar{o}^t_i)$ based on a single input $x_i$. Instead, it takes into account the amalgamation of information across relationships present in $\mathbb{O}_s$, $\mathbb{O}_t$, and $\mathbb{\bar{O}}_t$. 
	This implies that we incorporate more relationships into the loss function, depicted in Figure \ref{fig:SDC}. The solid line represents relationships used in prior works, while the dashed line signifies the relationship between distinct DNA fragments within a model.

	\begin{table*}[t] 
		\caption{Test accuracy of the proposed MGMP on CV task with and without (shown in parentheses) the DNA generator module.}\label{sample-table1full}
		\centering
		\begin{tabular}{ccccccc}
			\toprule 
			\multicolumn{2}{c}{Source}  & \multicolumn{2}{c}{Model pool} & \multicolumn{2}{c}{Evaluation}  & Performance  \\
			\cmidrule(r){1-2}  \cmidrule(r){3-4} \cmidrule(r){5-6} \cmidrule(r){7-7}
			Data      & Model       & Data      & Model  & Data   & Model  & Accuracy\\
			\midrule
			CIFAR10 & ResNet18  & CIFAR100 (3 6 8 9 5) &  ResNet18 & CIFAR100 (1 2 4 7 0) &  ResNet18 & 0.9193$\pm$ 0.0232 (0.7529$\pm$0.0347)  \\
			CIFAR10 & ResNet18  & CIFAR100 (6 7 0 1 2) &  ResNet18 & CIFAR100 (3 4 5 8 9) &  ResNet18 & 0.9511$\pm$ 0.0432 (0.8219$\pm$0.0332)  \\
			CIFAR10 & ResNet18  & CIFAR100 (9 3 7 8 1) &  ResNet18 & CIFAR100 (4 2 5 0 6) &  ResNet18 & 0.9607$\pm$ 0.0483 (0.8030$\pm$ 0.0451)  \\
			\midrule
			CIFAR10 & AlexNet  & CIFAR100 (3 6 8 9 5) &  AlexNet & CIFAR100 (1 2 4 7 0) &  AlexNet & 0.9025$\pm$ 0.0472 (0.7220$\pm$0.0687 )  \\ 
			CIFAR10 & AlexNet  & CIFAR100 (6 7 0 1 2) &  AlexNet & CIFAR100 (3 4 5 8 9) &  AlexNet & 0.9247$\pm$ 0.0392 (0.8471$\pm$0.0417)  \\
			CIFAR10 & AlexNet  & CIFAR100 (9 3 7 8 1) &  AlexNet & CIFAR100 (4 2 5 0 6) &  AlexNet & 0.9106$\pm$ 0.0331 (0.8231$\pm$0.0442)  \\
			\midrule
			CIFAR10 & ViT-small  & CIFAR100 (3 6 8 9 5) &  ViT-small & CIFAR100 (1 2 4 7 0) &  ViT-small & 0.9164$\pm$0.0224 (0.8326$\pm$0.0306)  \\
			\bottomrule
		\end{tabular}
	\end{table*}

	\begin{table*}[t] \scriptsize
		\centering
		\caption{Test accuracy of the proposed MGMP on NLP task with and without (shown in parentheses) the DNA generator module.}\label{sample-tablefull11}
		\begin{tabular}{ccccccc}
			\toprule 
			\multicolumn{2}{c}{Source}  & \multicolumn{2}{c}{Model pool} & \multicolumn{2}{c}{Evaluation}  & Performance  \\
			\cmidrule(r){1-2}  \cmidrule(r){3-4} \cmidrule(r){5-6} \cmidrule(r){7-7}
			Data     & Model       &  Data     & Model  &  Data  & Model & Accuracy\\
			\midrule
			AGNews & DistilBERT  & Amazon / DBPedia/YahooQA  &  DistilBERT & 20Newsgroups/Yelp &  DistilBERT & 0.8377$\pm$0.0778 (0.7203$\pm$0.0936)  \\
			YahooQA & DistilBERT  & 20Newsgroup/DBPedia/Amazon &  DistilBERT & AGNews/Yelp &  DistilBERT & 0.8625$\pm$0.0903 (0.7875$\pm$0.1003)  \\
			20Newsgroups & DistilBERT & YahooQA/Yelp/20Newsgroups &  DistilBERT & AGNews/DBPedia &  DistilBERT & 0.9395$\pm$0.0554 (0.8371$\pm$0.0964)  \\
			\midrule
			AGNews & BERT-base  & Amazon / DBPedia/YahooQA  &  BERT-base & 20Newsgroup/Yelp &  BERT-base & 0.8656 $\pm$0.0802 (0.7074$\pm$0.0918)  \\
			\bottomrule
		\end{tabular}
	\end{table*}

	Let $sim(u, v) = u^T v / ||u||||v||$ denote the dot product between $l_2$ normalized vectors $u$ and $v$ (i.e. cosine similarity).
	We consider a training dataset of $N$ examples and define a contrastive DNA generation loss on pairs of $(o^s_i, o^t_i)$ and $(o^s_i, \bar{o}^t_i)$. The loss function is as follows: 
	\begin{equation}
		\mathcal{L}_{S}=-\sum_{i=1}^{N}\log \frac{\exp(sim(o^s_i, o^t_i)/\tau)}{\sum_{k=1}^{N}\exp(sim(o^s_i, \bar{o}^t_k)/\tau) }
	\end{equation}
	where $\tau$ denotes a temperature parameter \cite{DBLP:conf/icml/ChenK0H20}. 
	
	We also consider the DNA distance in each model side:
	\begin{equation}
		\begin{aligned}
			\mathcal{L}_I =& -\sum_{i=1}^{N}\sum_{k=1}^{N-1} \log \big( \exp(sim(o^s_i, o^s_k)/\tau) \\
			& + \exp(sim(o^t_i, o^t_k)/\tau) \\
			& + \exp(sim(\bar{o}^t_i, \bar{o}^t_k)/\tau) \big), \ \ \  i\neq k.
		\end{aligned}
	\end{equation}

	The DNA similarity loss is then computed using
	\begin{equation}\label{danloss}
		\mathcal{L} = \mathcal{L}_{s} + \mathcal{L}_{I} + \lambda ||w_g||_2
	\end{equation}
	Here, $\mathcal{L}_{s}$ quantifies the disparity between the DNA of the source model and the target model, adhering to the first and second properties of the DNA latent space. On the other hand, $\mathcal{L}_{I}$ assesses the DNA stemming from the same model, aligning with the third property of the DNA latent space. The final term corresponds to $L_2$ regularization applied to the model parameters $w_g$ of the DNA generator.
	
	\vspace{1mm}
	\textit{Remark}: 
	In practice, after assembling the DNA fragments, we consider a cosine distance in our final decision (Eqn. (5) and (6)). Cosine similarity is the cosine of the angle between the vectors; that is, it is the dot product of the vectors divided by the product of their lengths. It follows that the cosine similarity does not depend on the magnitudes of the vectors, but only on their angle. Therefore, through DNA assemble with addition, the angles between the representations of the source and target models change even though the vectors only linearly move. That is why the generator can improve the results. Meanwhile, we emphasize that the part of DNA assemble is flexible, which means it can be replaced by different combining methods, e.g., concatenation or adding some networks (e.g., fully connected layers).

	\subsubsection{Provenance classifier}
	The outcome prediction network (i.e., a classifier) is employed to estimate the results of provenance prediction $h'$ by taking DNA representations as input. Let $f_p(\cdot)$ denote the function learned by the outcome prediction network. We concatenate $o^s_i$ and $o^t_i$ (or $o^s_i$ and $\bar{o}^t_i$) as input $o_i^c$.
	The loss function is as follows:
	\begin{equation}\label{pred}
		\mathcal{L}_{BCE} = -\frac{1}{N}\sum_{i=1}^N \left[h_i\log(h'_i)+(1-h_i)\log(1-h'_i)\right]
	\end{equation}
	where $h_i$ is the truth label on each input $o_i^c$ and $h_i' \leftarrow f_p(o_i^c)$. Note that $h = 1$ if $M_s$ and $M_t$ are homologous and $0$ otherwise.

	\subsubsection{Joint optimization}
	
	Both the DNA generation and the outcome prediction network are conventional feed-forward neural networks, enriched with Dropout \cite{DBLP:journals/jmlr/SrivastavaHKSS14} and the Rectified Linear Unit (ReLU) activation function. The global optimization problem is tackled by jointly optimizing the overall loss functions described in Eqn. (\ref{danloss}) and (\ref{pred}). Adam \cite{DBLP:journals/corr/KingmaB14} is adopted to solve the optimization problem.

	\subsubsection{Prediction }
	In the prediction phase, our framework is equipped to provide predictions at various levels of granularity concerning the data. Specifically, we can choose any model (e.g., $M_t$) to ascertain its homology with the source model $M_s$, based on either the DNA fragment $o$ or the DNA set $\mathbb{O}$. For predictions at the level of DNA fragments, the process is carried out as follows: 
	\begin{eqnarray}\label{vf}
		f(x_i,M_s,M_t)=
		\begin{cases} 1, & \text{if} \ f_p(o^c_i) \rightarrow 1\\
			0, &  \text{otherwise}
		\end{cases}
	\end{eqnarray}
	where $o^c_i$ is the concatenation of DNA fragment $o^s_i$ and $o^t_i$ corresponding to $x_i$ for $x_i \in D_s$. For the granularity of the model DNA set, we can compute the average results by considering the whole DNA fragments: $f(D_s,M_s,M_t)=1$ if $\frac{1}{N}\sum_{i = 1}^N f_p(o^c_i) \rightarrow 1$, and $0$ otherwise. In practice,  we can also employ a threshold $\delta$ for determining the final prediction, i.e., $\frac{1}{N}\sum_{i = 1}^N f_p(o^c_i) \geq \delta$. and $\delta$ can be seen as a degree of similarity of DNA. The choice of threshold depends on the specific context and requirements of the problem at hand.

	\section{Experiments}
	\label{sec:exp}
	To evaluate the effectiveness of our proposed framework, we conduct experiments on several commonly used benchmark datasets in Computer Vision and Natural Language Processing.

	\begin{table}[t]
		\centering
		\caption{The details of MGMP}
		\begin{tabular}{cccc}
			\toprule 
			Task & Generator &  |$z$| & |$o$|\\ 
			\hline
			CV & ResNet50 & 10 & 20 \\
			NLP & DistilBERT & 768 & 1536 \\
			\bottomrule
		\end{tabular}
		\label{PA}
	\end{table}
	
	\subsection{Experimental settings.}

	All experiments were conducted using Python programming language on a machine equipped with an Intel Core CPU, 64 GB of memory, and an NVIDIA Corporation GV100GL [Tesla V100 SXM2 32GB] GPU. We show the details of our proposed framework MGMP in Table \ref{PA}. 
	In the Computer Vision (CV) task, the generator's structure is set using the ResNet50 architecture. ResNet50 is a popular Convolutional Neural Network (CNN) architecture commonly used for image classification tasks. It consists of 50 layers, including convolutional layers, pooling layers, and fully connected layers. The model parameters, such as the learning rate, are set to their default values. In the NLP task, the structure of the generator is set as  DistilBERT, and the model parameters are also set to their default values. We employ three fully connected layers as the classifier in each task. Details regarding the impact of various generators can be found in Section \ref{amgmp}.

	\begin{table}[t]
		\centering
		\caption{Description of Datasets}
		\label{tab:datasets}
		\begin{tabular}{ccc}
			\toprule 
			\textbf{Dataset Name} & \textbf{Number of Samples} & \textbf{Features} \\
			\hline
			CIFAR-10  & 50,000 & RGB (32$\times$32)  \\
			\hline
			CIFAR-100   & 50,000 & RGB (32$\times$32)   \\
			\hline
			AGNews  & 20,000 & Text (4 classes)  \\
			\hline
			YahooQA  & 20,000 & Text (10 classes) \\
			\hline
			20 Newsgroups  & 13,370 & Text (20 classes)\\
			\hline
			DBPedia  & 20,000 & Text (14 classes)\\
			\hline
			Amazon Reviews  & 20,000 & Text (5 classes)\\
			\hline
			Yelp Review & 20,000 & Text (5 classes) \\
			\bottomrule
		\end{tabular}
	\end{table}

	Table \ref{tab:datasets} summarizes the datasets used in this study, including the dataset name, a brief description, the number of samples, and the types of features in each dataset. To accommodate the computational requirements and ensure efficient experimentation, we sampled a subset (20,000) from some NLP datasets as the training data.

	\subsection{Evaluation on CV tasks}
	
	\textbf{Setup.} In the CV experiment, we focus on a common image classification task. Here's an example to illustrate the process: we start by initializing the model architecture, such as ResNet18 \cite{DBLP:conf/cvpr/HeZRS16}, and then train the source model $M_s$ using CIFAR10. Then we utilize CIFAR100 to create model pools and evaluation datasets. To achieve this, we partition CIFAR100 into 10 disjoint 10-way classification subsets. Out of these, we randomly select 5 subsets (e.g., 3, 6, 8, 9, 5) to train 10 models (consisting of 5 homologous and 5 non-homologous models) that become part of the model pool. The remaining subsets are used to train multiple homologous and non-homologous models, which then serve as evaluation data. This setup is further detailed in the first row of Table \ref{sample-table1full}.
	
	\noindent\textbf{Experimental results.} 
	Table \ref{sample-table1full} presents the results of our proposed method under ResNet18, AlexNet, and ViT-small model structures on various datasets. We show performance on the granularity of the DNA fragment. Let $N$ be the total number of training data of the source model and $A_n$ ($A_o$) be the total number of data that can be identified correctly if we test the homologous (non-homologous) model as the target model. The test accuracy is defined as $Accuracy = \frac{A_{n}+A_{o}}{2N}$. Our method's performance is consistently superior across all datasets, demonstrating its effectiveness in verifying the provenance of models. 
	
	To the best of our knowledge, no method has been designed for the MP task. We use MGMP with and without the DNA generator module (Figure~\ref{fig:my_model}) as a baseline method and observe that the inclusion of the DNA generator improves the provenance prediction accuracy. Specifically, the test accuracy obtained without the DNA generator (shown in parentheses) is lower, highlighting the importance of the DNA generator module in our proposed framework.

	To provide a thorough insight into how DNA fragments contribute to the overall prediction, we analyze their influence by assessing performance at the DNA fragment level.
	Additionally, for assessing performance in terms of the DNA level, a threshold $\delta$ (e.g., 0.9) can be utilized to make final predictions. As demonstrated in Table 1, we consistently achieve the right provenance prediction in each row.

	\subsection{Evaluation on NLP tasks}
	
	\textbf{Setup.} For text classification tasks on various benchmark NLP datasets, we utilize the DistilBERT and BERT-base architectures as shown in Table \ref{sample-tablefull11}.  Similar to the experimental setup for CV tasks, we randomly select one dataset to train the source model, and three datasets to form the model pool. For evaluation, we use two datasets to train homologous or non-homologous models.

	\begin{figure*}[t]
		\centering
		\subfigure[]{\includegraphics[width=3.3cm, height=3.3cm]{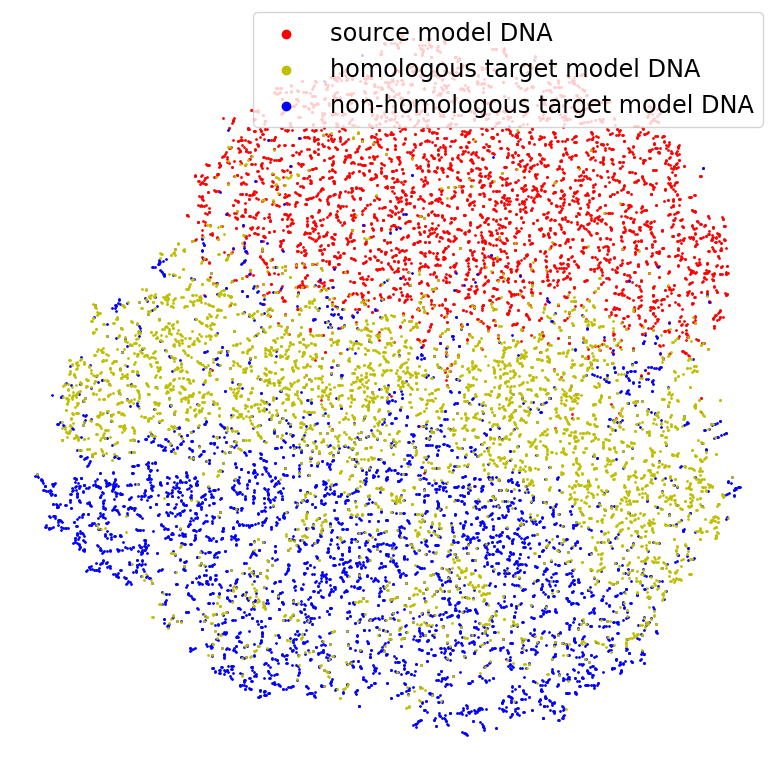}} \ \ 
		\subfigure[]{\includegraphics[width=3.3cm, height=3.3cm]{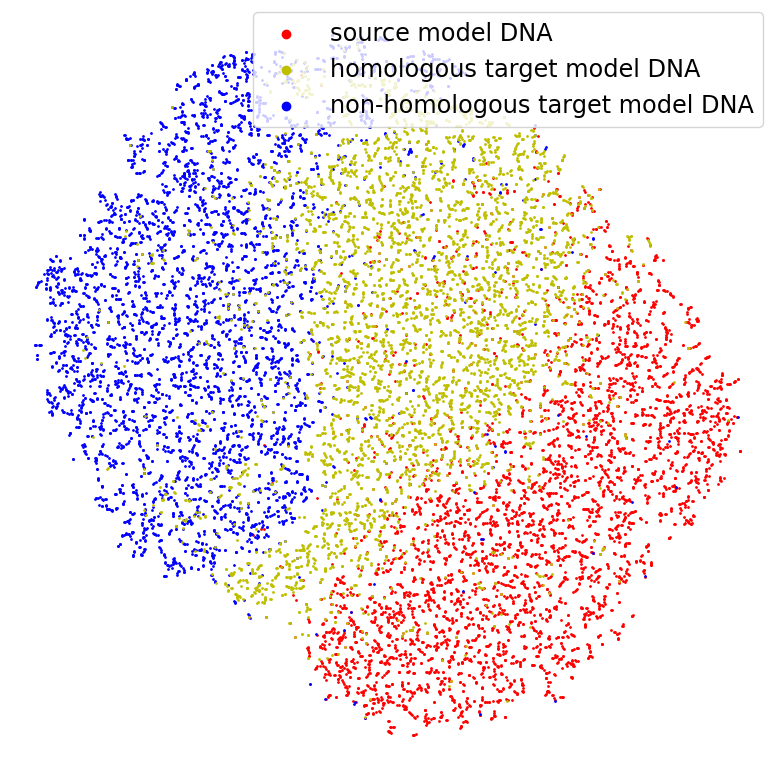}} \ \ 
		\subfigure[]{\includegraphics[width=3.3cm, height=3.3cm]{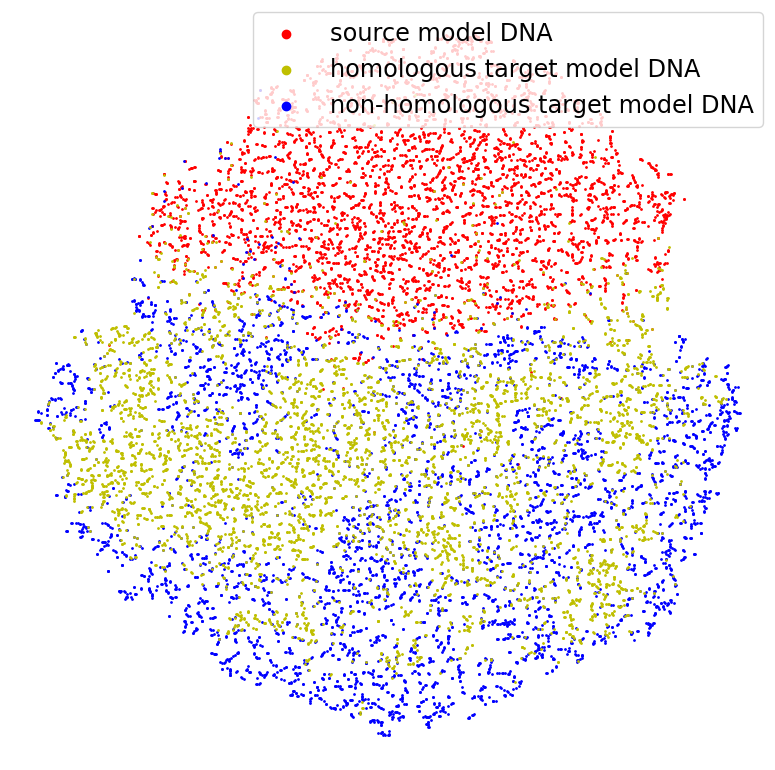}} \ \ 
		\subfigure[]{\includegraphics[width=3.3cm, height=3.3cm]{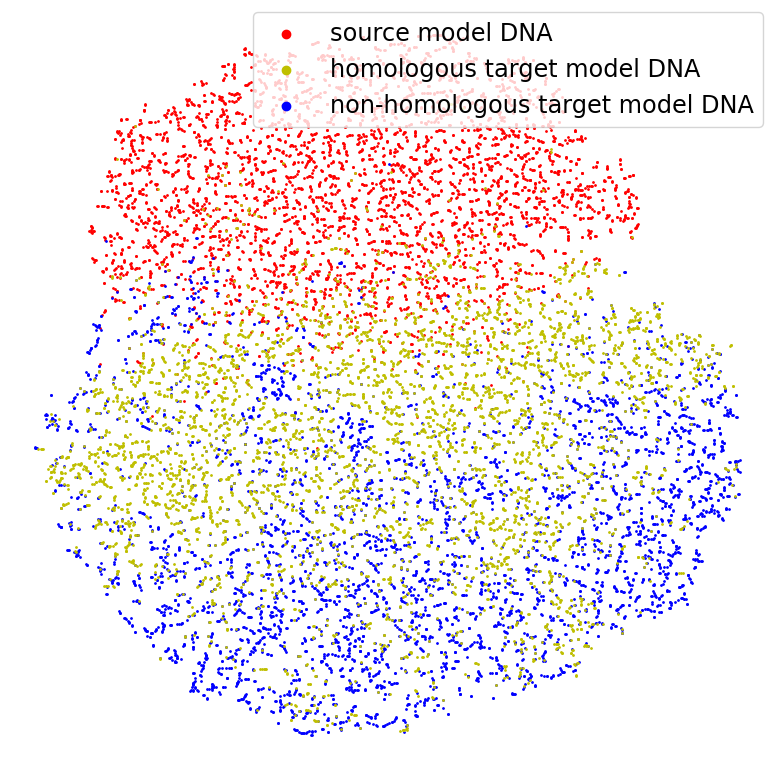}} \ \ 
		\subfigure[]{\includegraphics[width=3.3cm, height=3.3cm]{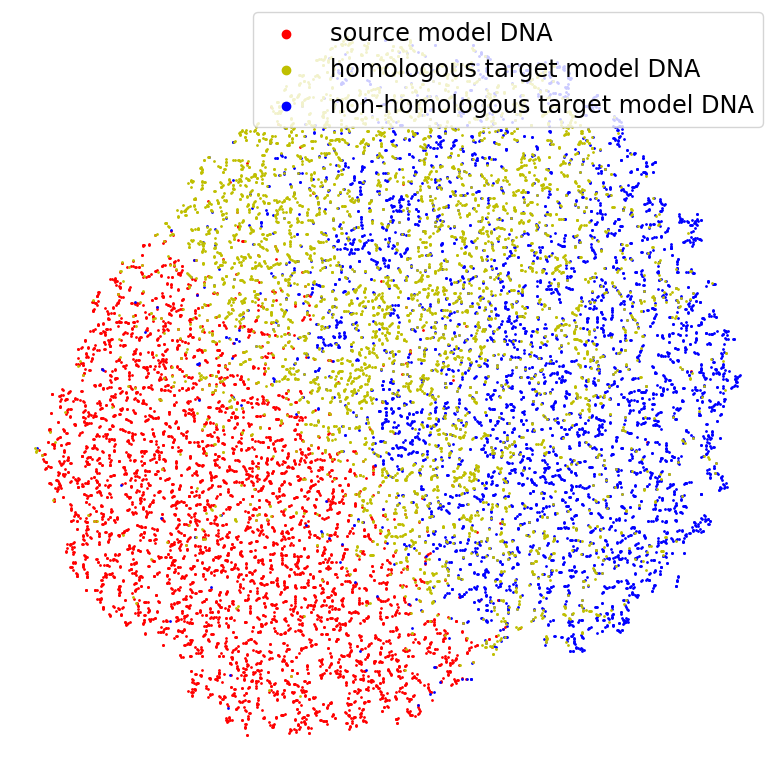}} \\
		\subfigure[]{\includegraphics[width=3.3cm, height=3.3cm]{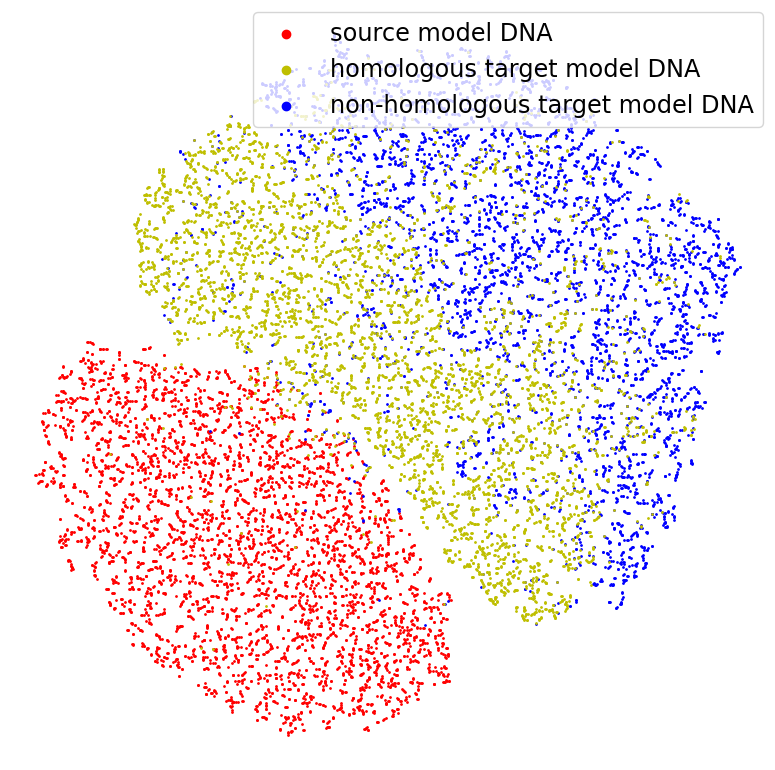}} \ \ 
		\subfigure[]{\includegraphics[width=3.3cm, height=3.3cm]{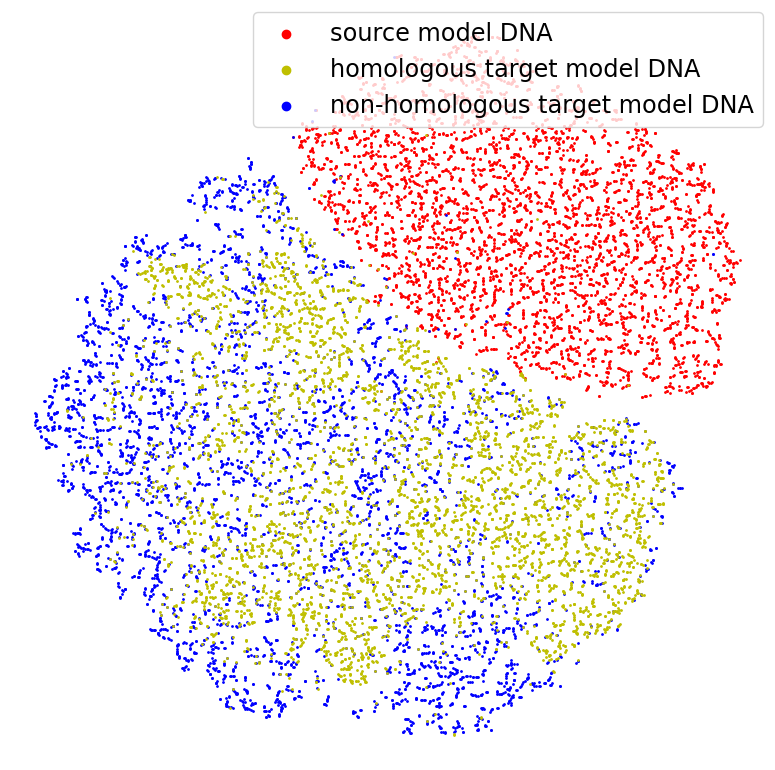}} \ \ 
		\subfigure[]{\includegraphics[width=3.3cm, height=3.3cm]{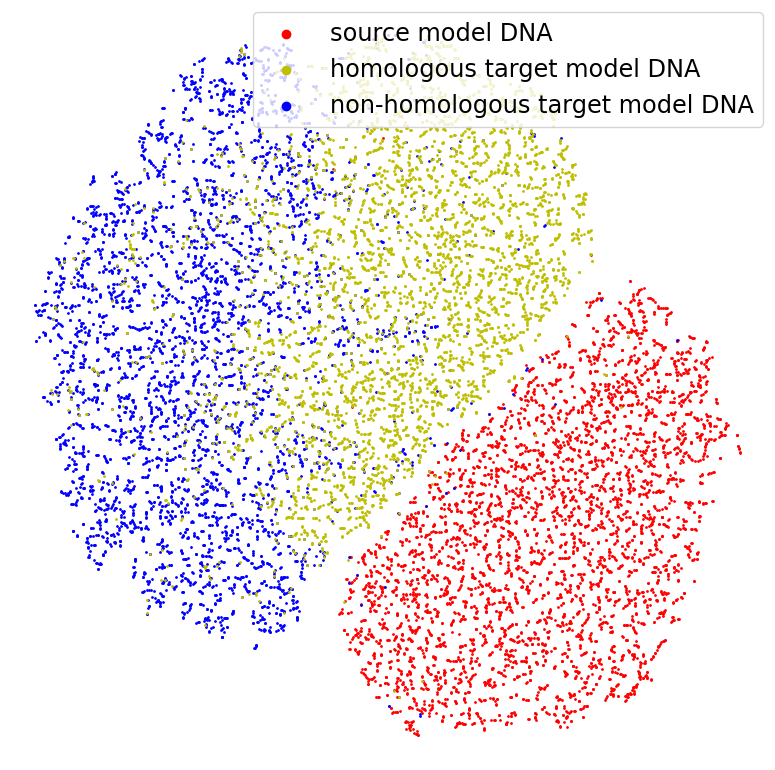}}  
		\subfigure[]{\includegraphics[width=3.3cm, height=3.3cm]{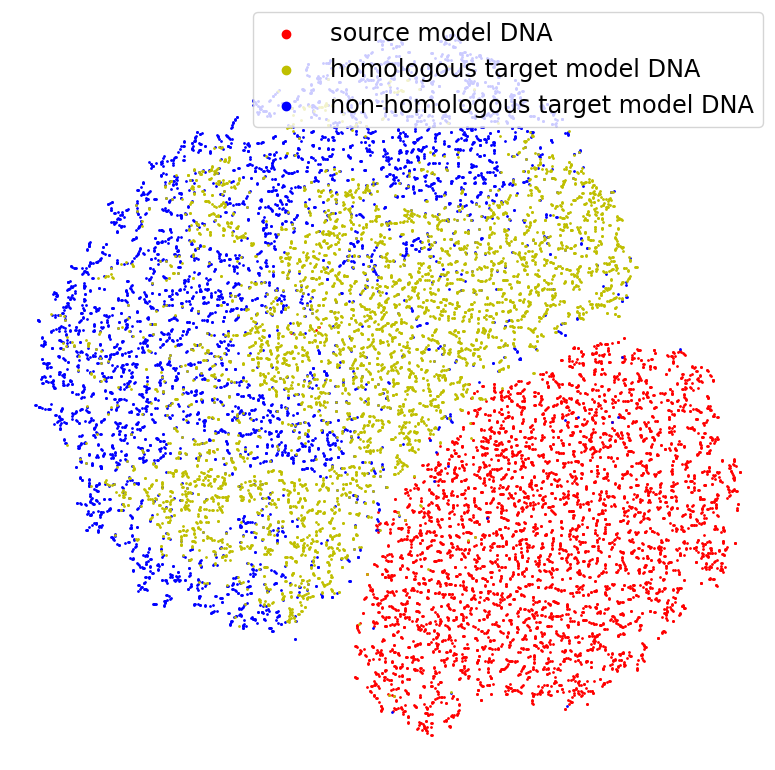}} \ \ 
		\subfigure[]{\includegraphics[width=3.3cm, height=3.3cm]{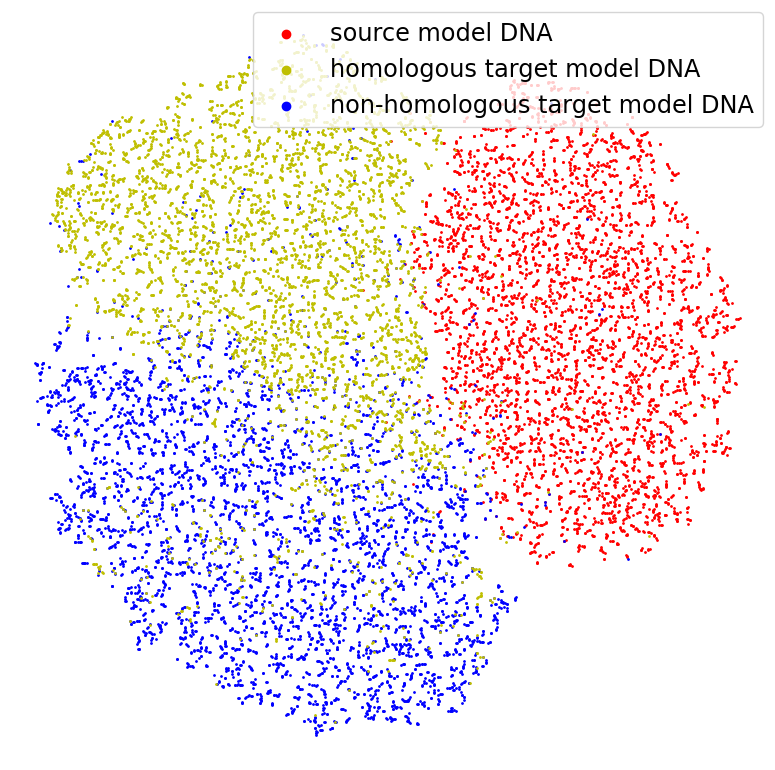}}
		\caption{The visualization of the DNA fragments of the source model (red), homologous target model (yellow), and non-homologous target model (blue).}
		\label{fig:vi}
	\end{figure*}

	\noindent\textbf{Experimental results.} 
	The results in Table \ref{sample-tablefull11} demonstrate that MGMP performs exceptionally well on the text classification task when applied to the DistilBERT model. 
	Similarly, MGMP without the DNA generator achieves worse accuracy than the original MGMP. These results are consistent with previous evaluations on CV task. The consistent findings across both the text classification and computer vision tasks emphasize the robustness and generalizability of MGMP.

	\subsection{Experimental visualization.} 
	We visualize the generated DNA fragments of the first CV experiment (i.e., the first row of Table \ref{sample-table1full}). Figure \ref{fig:vi} shows T-SNE visualizations of the space of DNA fragments results. The red points represent the generated DNA of the source model, the yellow ones are DNA from homologous target models, and the blue ones are DNA from non-homologous target models. We observe that the points belonging to homologous models are almost much closer to each other than points from non-homologous models. Specifically, the distance between the red and yellow points is much smaller than the distance between the red and blue points. This suggests that our framework is effective in capturing the similarity between homologous models and distinguishing them from non-homologous models.

	\subsection{Ablation study}\label{amgmp}
	The analysis of the MGMP framework aims to gain a comprehensive understanding of the individual components and processes within the approach. In this study, we specifically investigate the influence of different generator structures and various methods of DNA assembly. To isolate the effects of these factors, we conduct separate evaluations where one component is varied while keeping other components fixed. To establish a baseline experimental set, we consider the first row of Table \ref{sample-table1full} as the initial configuration.

	We observed that the performance of different generator structures remained consistent across various evaluation metrics. We found that the DNA assembly process had a notable effect on the results. Different ways of assembling the DNA fragments resulted in variations in performance. Our results indicate that increasing the dimensionality of the DNA representation can lead to improved results within the MGMP framework. By incorporating additional dimensions into the DNA, we can capture more nuanced information and potentially enhance the performance of the model.
	By analyzing the results in Table \ref{parameters}, we gain insights into the contributions of individual components and their interactions within the MGMP framework. This analysis helps us identify the optimal configuration and provides valuable guidance for future improvements and refinements of the approach.

	\begin{table}[t]
		\centering
		\caption{Components analysis: Test accuracy of MGMP with various DNA generator structures.}
		\begin{tabular}{ccc}
			\toprule 
			\multicolumn{2}{c}{Component}  & 
			Performance  \\
			\cmidrule(r){1-2}  \cmidrule(r){3-3} 
			Generator & DNA assemble ($|o|$)    & Accuracy\\
			\midrule
			ResNet50 &  addition (10) & 0.9193 \\
			ViT-samll &  addition (10)  & 0.9004 \\
			ViT-base &  addition (10)  & 0.9137 \\
			ResNet50 &  concatenate (20) & 0.9264 \\
			ResNet50 &  concatenate (60) & 0.9416 \\
			ResNet50 &  concatenate (110) & 0.9478 \\
			\bottomrule
		\end{tabular}
		\label{parameters}
	\end{table}
	
	\section{Conclusion}
	In this paper, we present an efficient model representation learning framework for tackling an important problem, namely Model Provenance. We introduce a new idea of model DNA to represent a machine learning model. The proposed framework first constructs the model DNA space which preserves similarity information between homologous models and enhances the differences between non-homologous models, and then uses model DNA to obtain provenance outcomes. Experimental results on different tasks show that our method achieves good performance in the MP task.

	\bibliography{mybibfile}

\begin{thebibliography}{36}
\providecommand{\natexlab}[1]{#1}
\providecommand{\url}[1]{\texttt{#1}}
\expandafter\ifx\csname urlstyle\endcsname\relax
  \providecommand{\doi}[1]{doi: #1}\else
  \providecommand{\doi}{doi: \begingroup \urlstyle{rm}\Url}\fi

\bibitem[Aljundi et~al.(2017)Aljundi, Chakravarty, and
  Tuytelaars]{DBLP:conf/cvpr/AljundiCT17}
R.~Aljundi, P.~Chakravarty, and T.~Tuytelaars.
\newblock Expert gate: Lifelong learning with a network of experts.
\newblock In \emph{CVPR}, pages 7120--7129, 2017.

\bibitem[Bengio et~al.(2013)Bengio, Courville, and
  Vincent]{DBLP:journals/pami/BengioCV13}
Y.~Bengio, A.~C. Courville, and P.~Vincent.
\newblock Representation learning: {A} review and new perspectives.
\newblock \emph{{IEEE} Transactions on Pattern Analysis and Machine
  Intelligence.}, 35\penalty0 (8):\penalty0 1798--1828, 2013.

\bibitem[Chaudhry et~al.(2019)Chaudhry, Ranzato, Rohrbach, and
  Elhoseiny]{DBLP:conf/iclr/ChaudhryRRE19}
A.~Chaudhry, M.~Ranzato, M.~Rohrbach, and M.~Elhoseiny.
\newblock Efficient lifelong learning with {A-GEM}.
\newblock In \emph{ICLR}, 2019.

\bibitem[Chen et~al.(2020)Chen, Kornblith, Norouzi, and
  Hinton]{DBLP:conf/icml/ChenK0H20}
T.~Chen, S.~Kornblith, M.~Norouzi, and G.~E. Hinton.
\newblock A simple framework for contrastive learning of visual
  representations.
\newblock In \emph{ICML}, pages 1597--1607, 2020.

\bibitem[Chen and Liu(2018)]{DBLP:series/synthesis/2018Chen}
Z.~Chen and B.~Liu.
\newblock \emph{Lifelong Machine Learning, Second Edition}.
\newblock Synthesis Lectures on Artificial Intelligence and Machine Learning.
  Morgan {\&} Claypool Publishers, 2018.

\bibitem[Csisz{\'{a}}rik et~al.(2021)Csisz{\'{a}}rik,
  Kor{\"{o}}si{-}Szab{\'{o}}, Matszangosz, Papp, and
  Varga]{DBLP:conf/nips/CsiszarikKMPV21}
A.~Csisz{\'{a}}rik, P.~Kor{\"{o}}si{-}Szab{\'{o}}, {\'{A}}.~K. Matszangosz,
  G.~Papp, and D.~Varga.
\newblock Similarity and matching of neural network representations.
\newblock In \emph{NeurIPS}, pages 5656--5668, 2021.

\bibitem[Dixon et~al.(2020)Dixon, Halperin, and Bilokon]{dixon2020machine}
M.~F. Dixon, I.~Halperin, and P.~Bilokon.
\newblock \emph{Machine learning in Finance}, volume 1170.
\newblock Springer, 2020.

\bibitem[Goodfellow et~al.(2020)Goodfellow, Pouget{-}Abadie, Mirza, Xu,
  Warde{-}Farley, Ozair, Courville, and
  Bengio]{DBLP:journals/cacm/GoodfellowPMXWO20}
I.~J. Goodfellow, J.~Pouget{-}Abadie, M.~Mirza, B.~Xu, D.~Warde{-}Farley,
  S.~Ozair, A.~C. Courville, and Y.~Bengio.
\newblock Generative adversarial networks.
\newblock \emph{Communications of the ACM}, 63\penalty0 (11):\penalty0
  139--144, 2020.

\bibitem[He et~al.(2016)He, Zhang, Ren, and Sun]{DBLP:conf/cvpr/HeZRS16}
K.~He, X.~Zhang, S.~Ren, and J.~Sun.
\newblock Deep residual learning for image recognition.
\newblock In \emph{CVPR}, pages 770--778, 2016.

\bibitem[Hu et~al.(2021)Hu, Salcic, Dobbie, and
  Zhang]{DBLP:journals/corr/abs-2103-07853}
H.~Hu, Z.~Salcic, G.~Dobbie, and X.~Zhang.
\newblock Membership inference attacks on machine learning: {A} survey.
\newblock \emph{CoRR}, abs/2103.07853, 2021.

\bibitem[Keskar et~al.(2017)Keskar, Mudigere, Nocedal, Smelyanskiy, and
  Tang]{DBLP:conf/iclr/KeskarMNST17}
N.~S. Keskar, D.~Mudigere, J.~Nocedal, M.~Smelyanskiy, and P.~T.~P. Tang.
\newblock On large-batch training for deep learning: Generalization gap and
  sharp minima.
\newblock In \emph{ICLR}, 2017.

\bibitem[Kingma and Ba(2015)]{DBLP:journals/corr/KingmaB14}
D.~P. Kingma and J.~Ba.
\newblock Adam: {A} method for stochastic optimization.
\newblock In Y.~Bengio and Y.~LeCun, editors, \emph{ICLR}, 2015.

\bibitem[Kirkpatrick et~al.(2017)Kirkpatrick, Pascanu, Rabinowitz, Veness,
  Desjardins, Rusu, Milan, Quan, Ramalho, Grabska-Barwinska,
  et~al.]{kirkpatrick2017overcoming}
J.~Kirkpatrick, R.~Pascanu, N.~Rabinowitz, J.~Veness, G.~Desjardins, A.~A.
  Rusu, K.~Milan, J.~Quan, T.~Ramalho, A.~Grabska-Barwinska, et~al.
\newblock Overcoming catastrophic forgetting in neural networks.
\newblock \emph{Proceedings of the national academy of sciences}, 114\penalty0
  (13):\penalty0 3521--3526, 2017.

\bibitem[Kornblith et~al.(2019)Kornblith, Norouzi, Lee, and
  Hinton]{DBLP:conf/icml/Kornblith0LH19}
S.~Kornblith, M.~Norouzi, H.~Lee, and G.~E. Hinton.
\newblock Similarity of neural network representations revisited.
\newblock In \emph{ICML}, pages 3519--3529, 2019.

\bibitem[Krizhevsky et~al.(2012)Krizhevsky, Sutskever, and
  Hinton]{DBLP:conf/nips/KrizhevskySH12}
A.~Krizhevsky, I.~Sutskever, and G.~E. Hinton.
\newblock Imagenet classification with deep convolutional neural networks.
\newblock In \emph{NIPS}, pages 1106--1114, 2012.

\bibitem[Lange et~al.(2022)Lange, Aljundi, Masana, Parisot, Jia, Leonardis,
  Slabaugh, and Tuytelaars]{DBLP:journals/pami/LangeAMPJLST22}
M.~D. Lange, R.~Aljundi, M.~Masana, S.~Parisot, X.~Jia, A.~Leonardis, G.~G.
  Slabaugh, and T.~Tuytelaars.
\newblock A continual learning survey: Defying forgetting in classification
  tasks.
\newblock \emph{IEEE TPAMI}, 44\penalty0 (7):\penalty0 3366--3385, 2022.

\bibitem[Le{-}Khac et~al.(2020)Le{-}Khac, Healy, and
  Smeaton]{DBLP:journals/access/Le-KhacHS20}
P.~H. Le{-}Khac, G.~Healy, and A.~F. Smeaton.
\newblock Contrastive representation learning: {A} framework and review.
\newblock \emph{{IEEE} Access}, 8:\penalty0 193907--193934, 2020.

\bibitem[Lopez{-}Paz and Ranzato(2017)]{DBLP:conf/nips/Lopez-PazR17}
D.~Lopez{-}Paz and M.~Ranzato.
\newblock Gradient episodic memory for continual learning.
\newblock In \emph{NIPS}, pages 6467--6476, 2017.

\bibitem[Maini et~al.(2021)Maini, Yaghini, and
  Papernot]{DBLP:conf/iclr/MainiYP21}
P.~Maini, M.~Yaghini, and N.~Papernot.
\newblock Dataset inference: Ownership resolution in machine learning.
\newblock In \emph{ICLR}, 2021.

\bibitem[Mehta et~al.(2021)Mehta, Patil, Chandar, and
  Strubell]{DBLP:journals/corr/abs-2112-09153}
S.~V. Mehta, D.~Patil, S.~Chandar, and E.~Strubell.
\newblock An empirical investigation of the role of pre-training in lifelong
  learning.
\newblock \emph{CoRR}, abs/2112.09153, 2021.

\bibitem[Mirzadeh et~al.(2020)Mirzadeh, Farajtabar, Pascanu, and
  Ghasemzadeh]{DBLP:conf/nips/MirzadehFPG20}
S.~Mirzadeh, M.~Farajtabar, R.~Pascanu, and H.~Ghasemzadeh.
\newblock Understanding the role of training regimes in continual learning.
\newblock In \emph{NeurIPS}, 2020.

\bibitem[Mu et~al.(2022)Mu, Pang, and Zhu]{DBLP:journals/corr/abs-2209-01538}
X.~Mu, M.~Pang, and F.~Zhu.
\newblock Data provenance via differential auditing.
\newblock \emph{CoRR}, abs/2209.01538, 2022.

\bibitem[Parisi et~al.(2019)Parisi, Kemker, Part, Kanan, and
  Wermter]{DBLP:journals/nn/ParisiKPKW19}
G.~I. Parisi, R.~Kemker, J.~L. Part, C.~Kanan, and S.~Wermter.
\newblock Continual lifelong learning with neural networks: {A} review.
\newblock \emph{Neural Networks}, 113:\penalty0 54--71, 2019.

\bibitem[Paul et~al.(2021)Paul, Ganguli, and
  Dziugaite]{DBLP:conf/nips/PaulGD21}
M.~Paul, S.~Ganguli, and G.~K. Dziugaite.
\newblock Deep learning on a data diet: Finding important examples early in
  training.
\newblock In \emph{NeurIPS}, pages 20596--20607, 2021.

\bibitem[Ridley(2000)]{ridley2000genome}
M.~Ridley.
\newblock Genome: the autobiography of a species in 23 chapters.
\newblock \emph{Nature Medicine}, 6\penalty0 (1):\penalty0 11--11, 2000.

\bibitem[Rusu et~al.(2016)Rusu, Rabinowitz, Desjardins, Soyer, Kirkpatrick,
  Kavukcuoglu, Pascanu, and Hadsell]{DBLP:journals/corr/RusuRDSKKPH16}
A.~A. Rusu, N.~C. Rabinowitz, G.~Desjardins, H.~Soyer, J.~Kirkpatrick,
  K.~Kavukcuoglu, R.~Pascanu, and R.~Hadsell.
\newblock Progressive neural networks.
\newblock \emph{CoRR}, abs/1606.04671, 2016.

\bibitem[Shehab et~al.(2022)Shehab, Abualigah, Shambour, Abu{-}Hashem,
  Shambour, Alsalibi, and Gandomi]{DBLP:journals/cbm/ShehabASASAG22}
M.~Shehab, L.~Abualigah, Q.~Shambour, M.~A. Abu{-}Hashem, M.~K.~Y. Shambour,
  A.~I. Alsalibi, and A.~H. Gandomi.
\newblock Machine learning in medical applications: {A} review of
  state-of-the-art methods.
\newblock \emph{Computers in Biology and Medicine}, 145:\penalty0 105458, 2022.

\bibitem[Shokri et~al.(2017)Shokri, Stronati, Song, and
  Shmatikov]{DBLP:conf/sp/ShokriSSS17}
R.~Shokri, M.~Stronati, C.~Song, and V.~Shmatikov.
\newblock Membership inference attacks against machine learning models.
\newblock In \emph{S\&P}, pages 3--18, 2017.

\bibitem[Sodhani et~al.(2020)Sodhani, Chandar, and
  Bengio]{DBLP:journals/neco/SodhaniCB20}
S.~Sodhani, S.~Chandar, and Y.~Bengio.
\newblock Toward training recurrent neural networks for lifelong learning.
\newblock \emph{Neural Computing}, 32\penalty0 (1):\penalty0 1--35, 2020.

\bibitem[Song and Shmatikov(2019)]{DBLP:conf/kdd/SongS19}
C.~Song and V.~Shmatikov.
\newblock Auditing data provenance in text-generation models.
\newblock In \emph{KDD}, pages 196--206, 2019.

\bibitem[Srivastava et~al.(2014)Srivastava, Hinton, Krizhevsky, Sutskever, and
  Salakhutdinov]{DBLP:journals/jmlr/SrivastavaHKSS14}
N.~Srivastava, G.~E. Hinton, A.~Krizhevsky, I.~Sutskever, and R.~Salakhutdinov.
\newblock Dropout: a simple way to prevent neural networks from overfitting.
\newblock \emph{The Journal of Machine Learning Research}, 15\penalty0
  (1):\penalty0 1929--1958, 2014.

\bibitem[Toneva et~al.(2019)Toneva, Sordoni, des Combes, Trischler, Bengio, and
  Gordon]{DBLP:conf/iclr/TonevaSCTBG19}
M.~Toneva, A.~Sordoni, R.~T. des Combes, A.~Trischler, Y.~Bengio, and G.~J.
  Gordon.
\newblock An empirical study of example forgetting during deep neural network
  learning.
\newblock In \emph{ICLR}, 2019.

\bibitem[van~den Oord et~al.(2018)van~den Oord, Li, and
  Vinyals]{DBLP:journals/corr/abs-1807-03748}
A.~van~den Oord, Y.~Li, and O.~Vinyals.
\newblock Representation learning with contrastive predictive coding.
\newblock \emph{CoRR}, abs/1807.03748, 2018.

\bibitem[Wang et~al.(2020)Wang, Mehta, P{\'{o}}czos, and
  Carbonell]{DBLP:conf/emnlp/WangMPC20}
Z.~Wang, S.~V. Mehta, B.~P{\'{o}}czos, and J.~G. Carbonell.
\newblock Efficient meta lifelong-learning with limited memory.
\newblock In \emph{EMNLP}, pages 535--548, 2020.

\bibitem[Yuan et~al.(2021)Yuan, Lin, Kuen, Zhang, Wang, Maire, Kale, and
  Faieta]{DBLP:conf/cvpr/Yuan0K0WMKF21}
X.~Yuan, Z.~Lin, J.~Kuen, J.~Zhang, Y.~Wang, M.~Maire, A.~Kale, and B.~Faieta.
\newblock Multimodal contrastive training for visual representation learning.
\newblock In \emph{CVPR}, pages 6995--7004, 2021.

\bibitem[Zenke et~al.(2017)Zenke, Poole, and Ganguli]{DBLP:conf/icml/ZenkePG17}
F.~Zenke, B.~Poole, and S.~Ganguli.
\newblock Continual learning through synaptic intelligence.
\newblock In \emph{ICML}, pages 3987--3995, 2017.

\end{thebibliography}
	
	\clearpage
	
	\appendix
	
	\section{Discussion}
	
	\begin{figure*}[b]
		\centering
		\subfigure[]{\epsfig{file=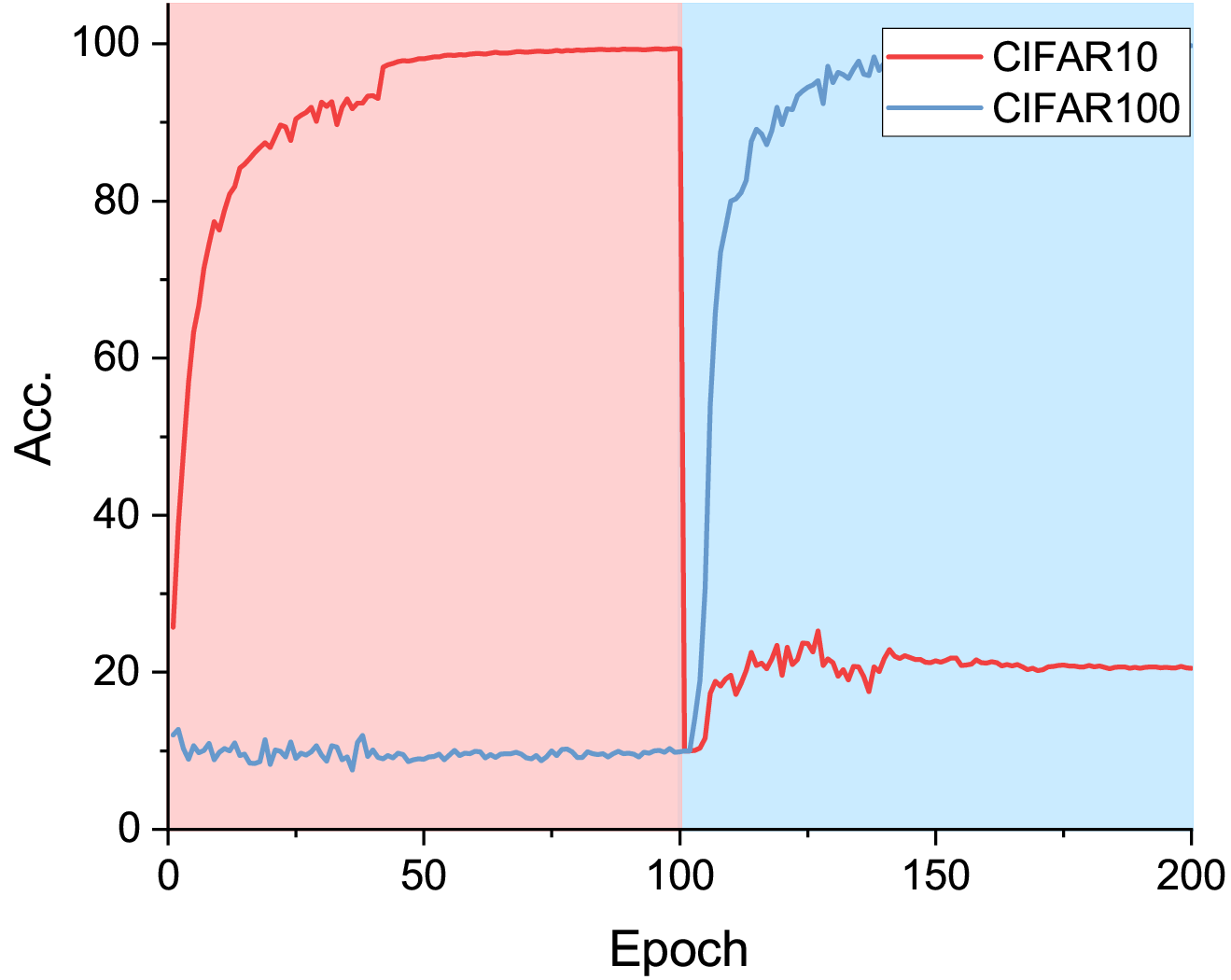,height=1in,width=0.9in}} 
		\subfigure[]{\epsfig{file=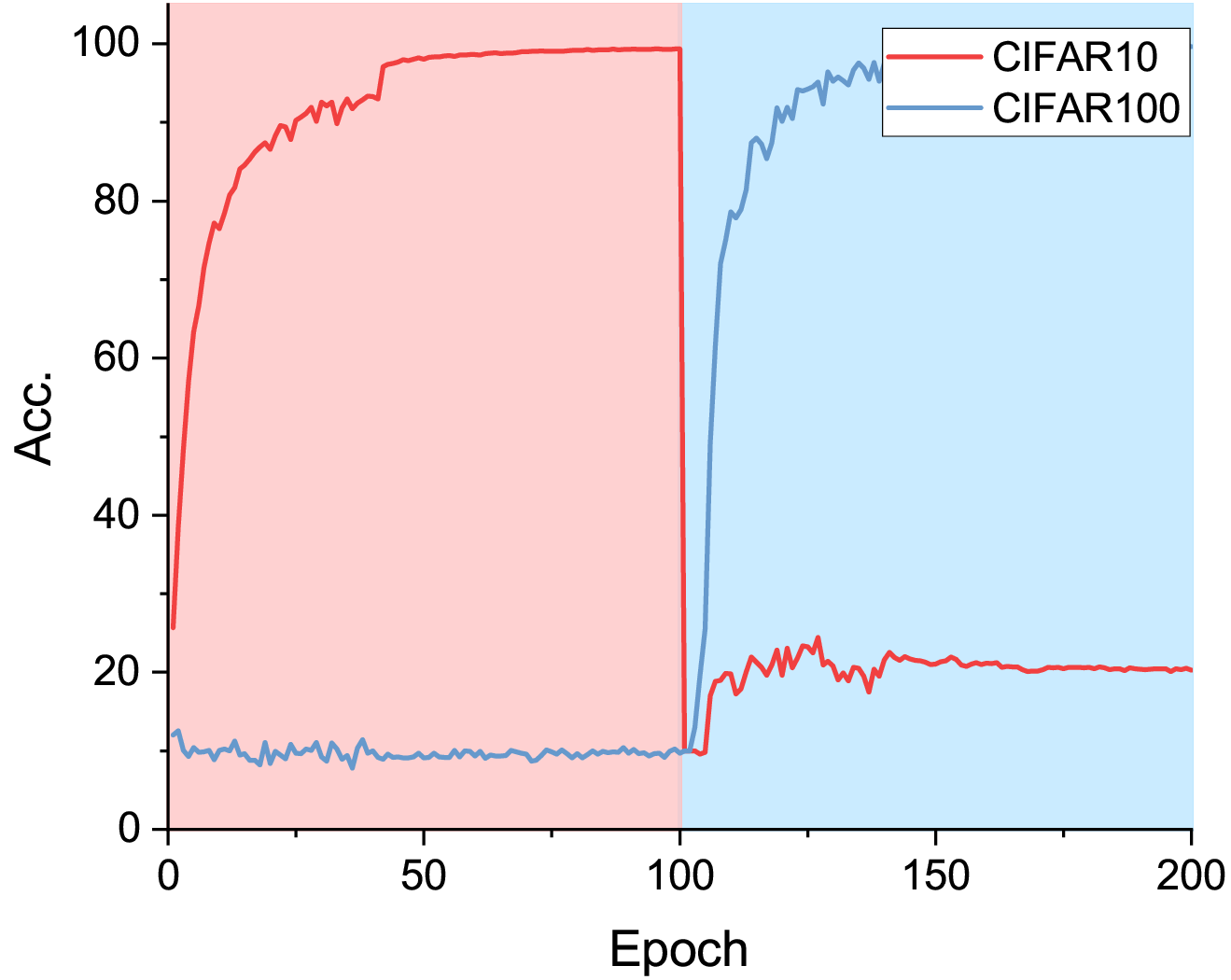,height=1in,width=0.9in}} 
		\subfigure[]{\epsfig{file=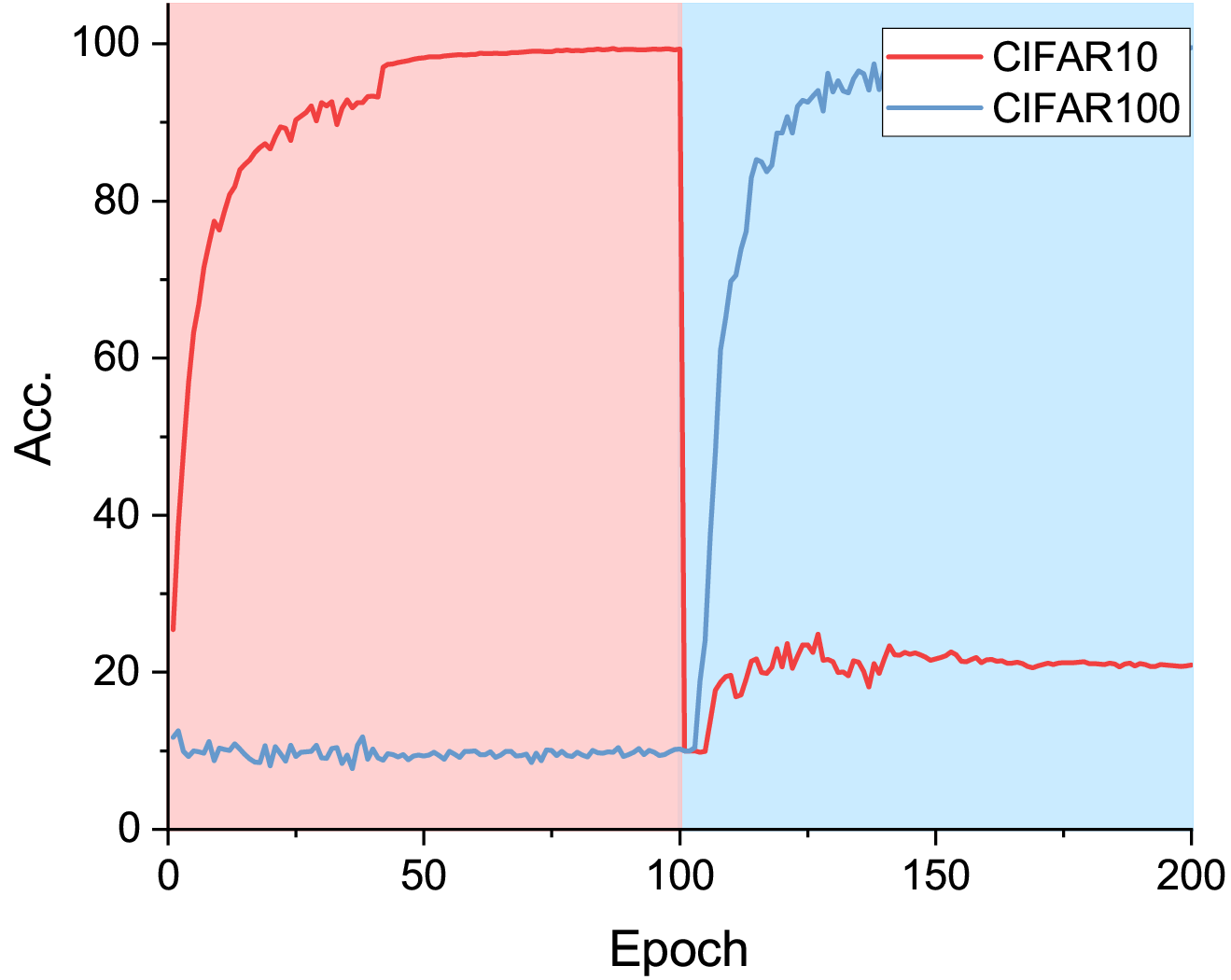,height=1in,width=0.9in}} 
		\subfigure[]{\epsfig{file=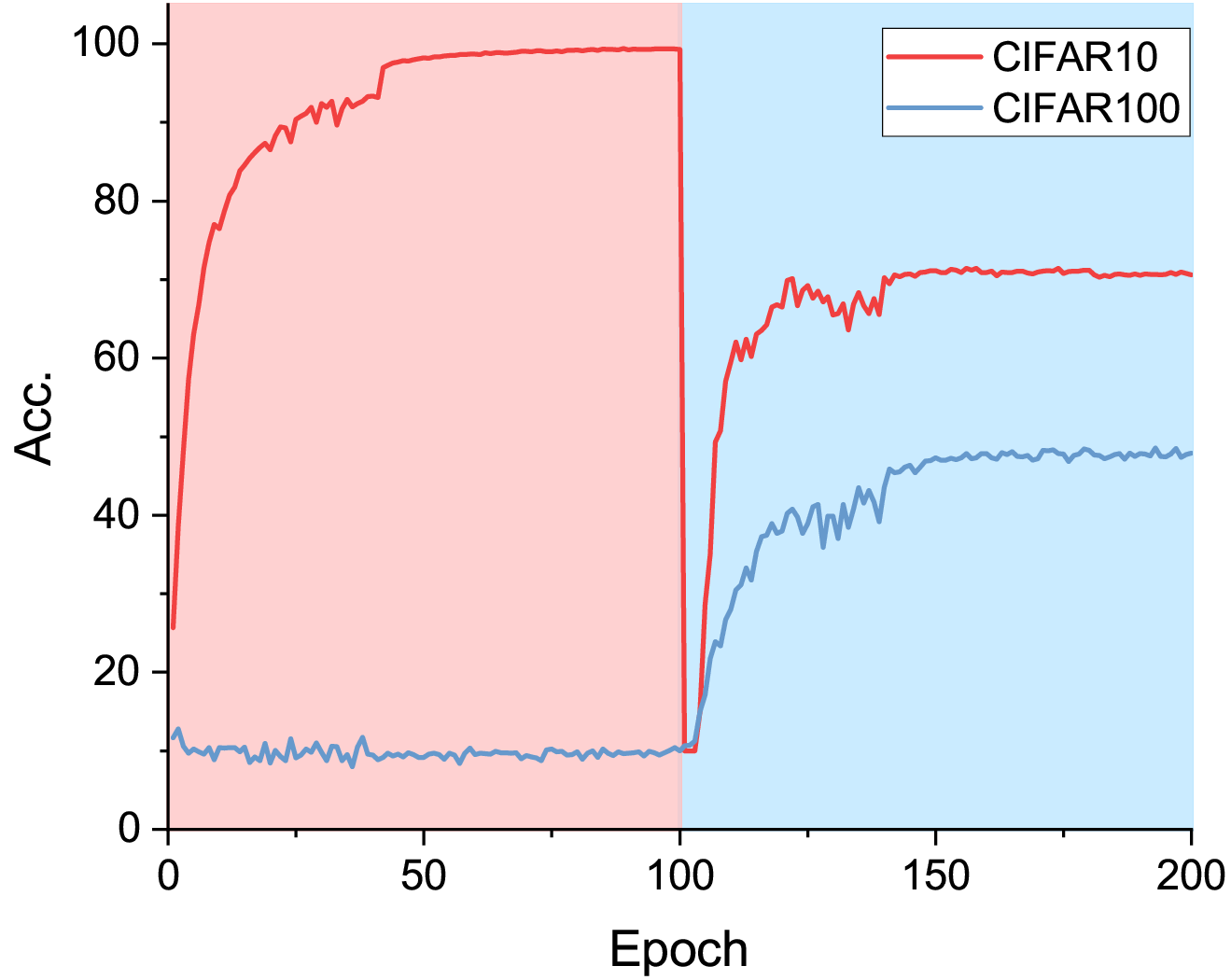,height=1in,width=0.9in}}  
		\subfigure[]{\epsfig{file=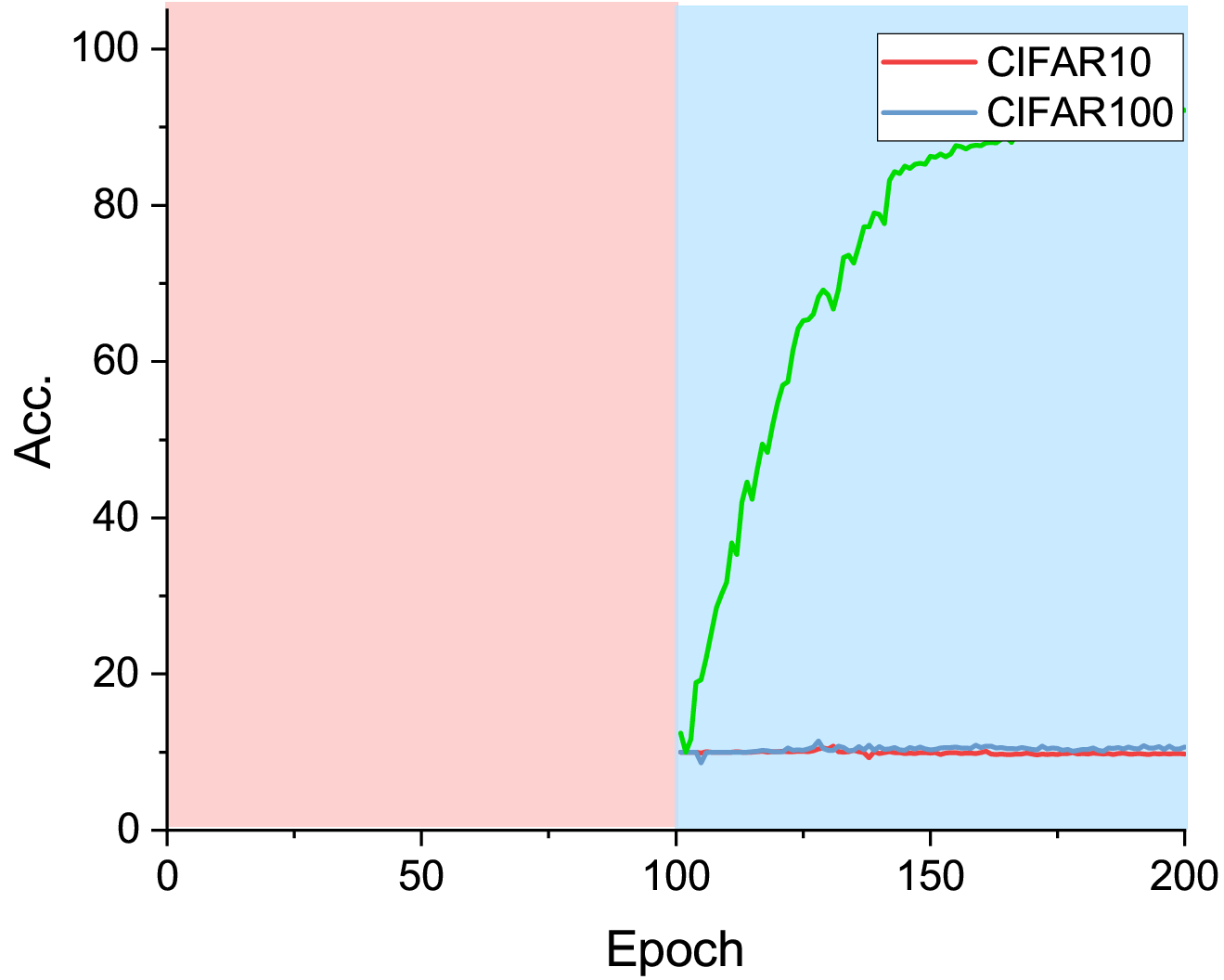,height=1in,width=0.9in}}  

		\caption{AlexNet. (a) No replace layer. (b) Replace target model's last layer with source model. (c) Replace last two layers. (d) Replace last three layers. (e) Replace target model (random initialization)'s last three layers with source model.}
		\label{fig:ALEXnet}
	\end{figure*}
	
	In addition to the previous ResNet experiments in Section \ref{sec:def}, we conducted another experiment on the model AlexNet, the results of which are presented in Figure \ref{fig:ALEXnet}. Remarkably, the findings from the AlexNet experiment align with the observations from the earlier experiments.

	\section{The experimental visualization on NLP task.}
	
	\begin{figure*}[htbp]
		\centering
		\subfigure[]{\includegraphics[width=2cm, height=2cm]{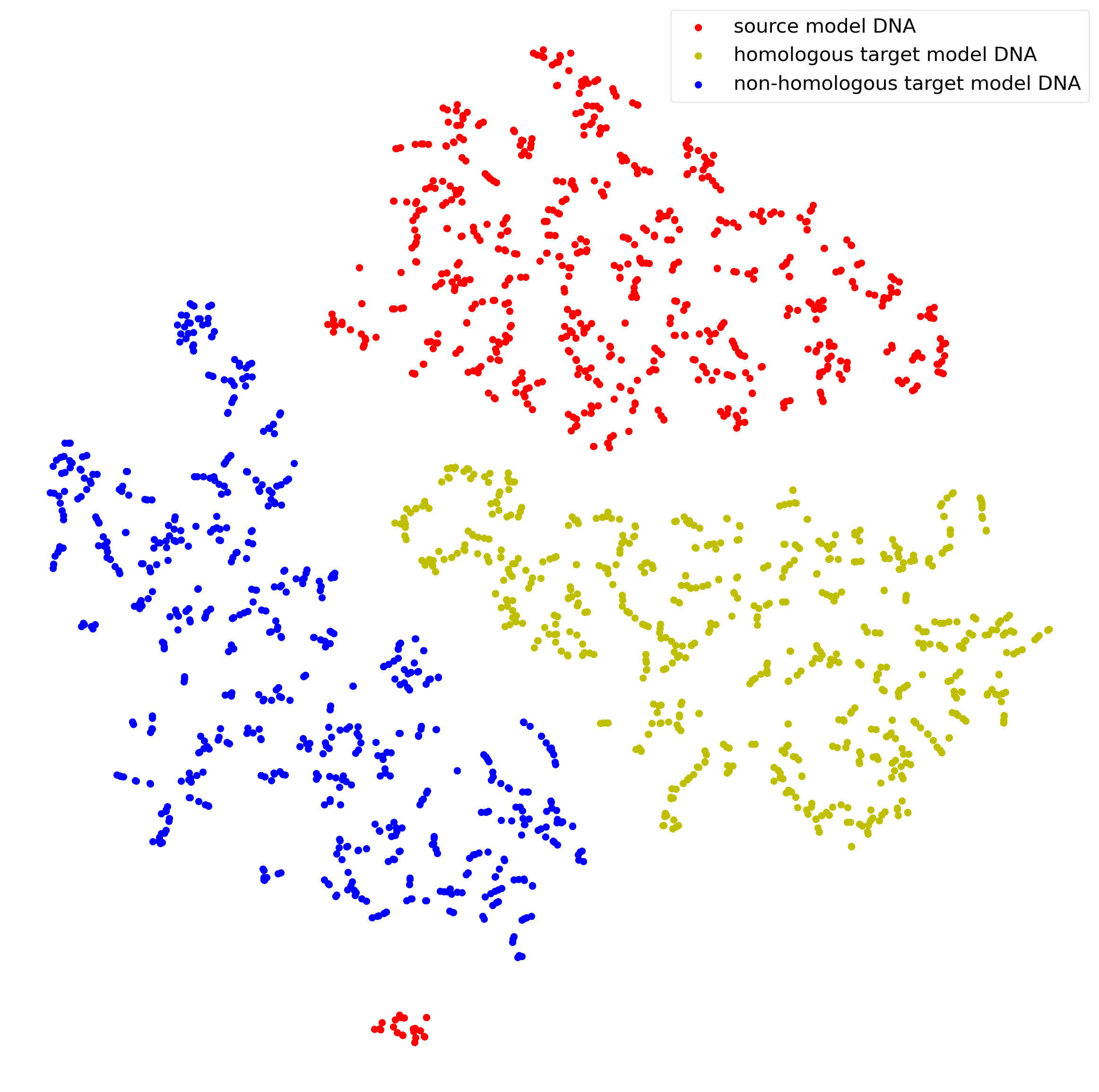}} \ \ 
		\subfigure[]{\includegraphics[width=2cm, height=2cm]{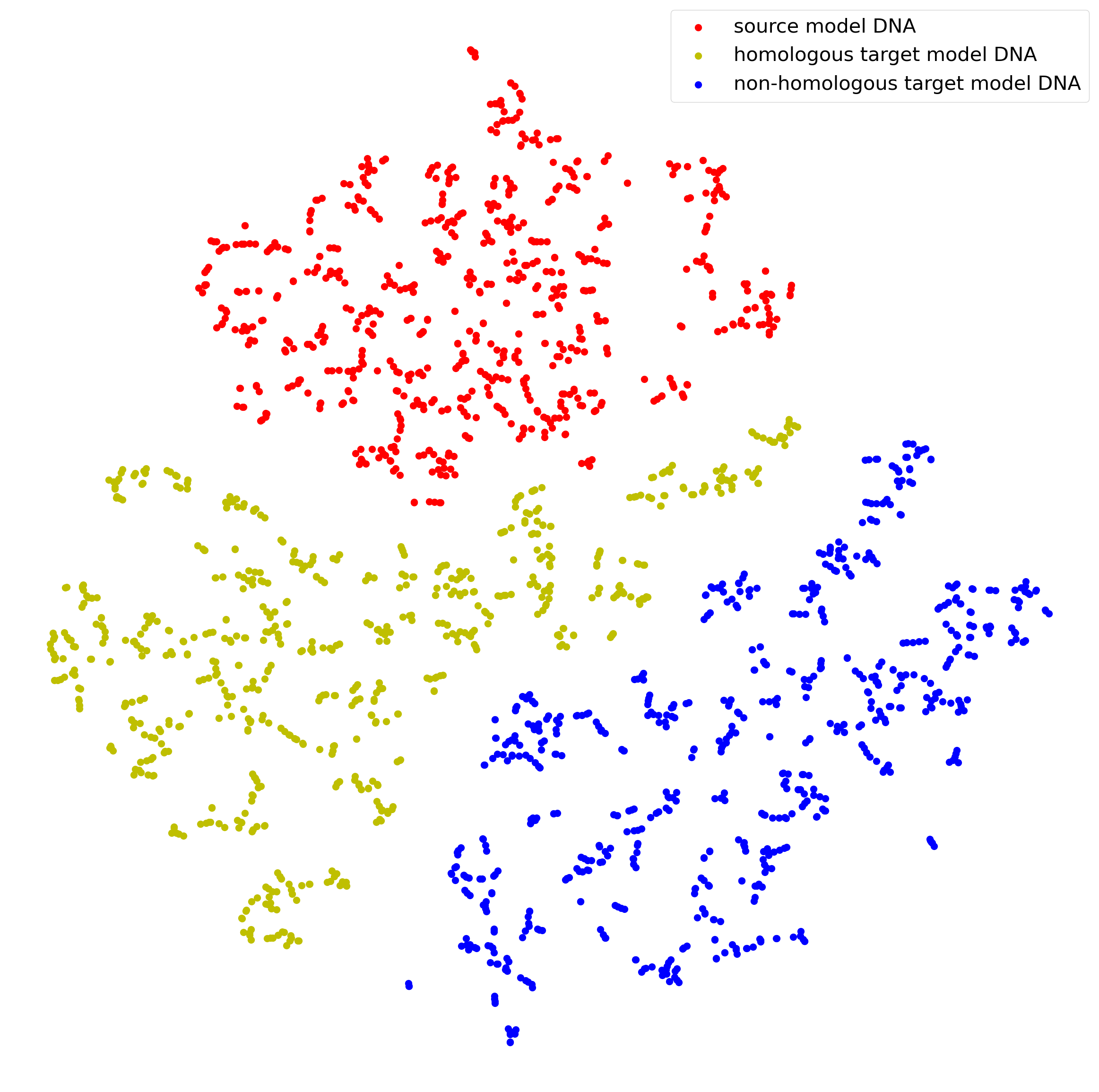}} \ \ 
		\subfigure[]{\includegraphics[width=2cm, height=2cm]{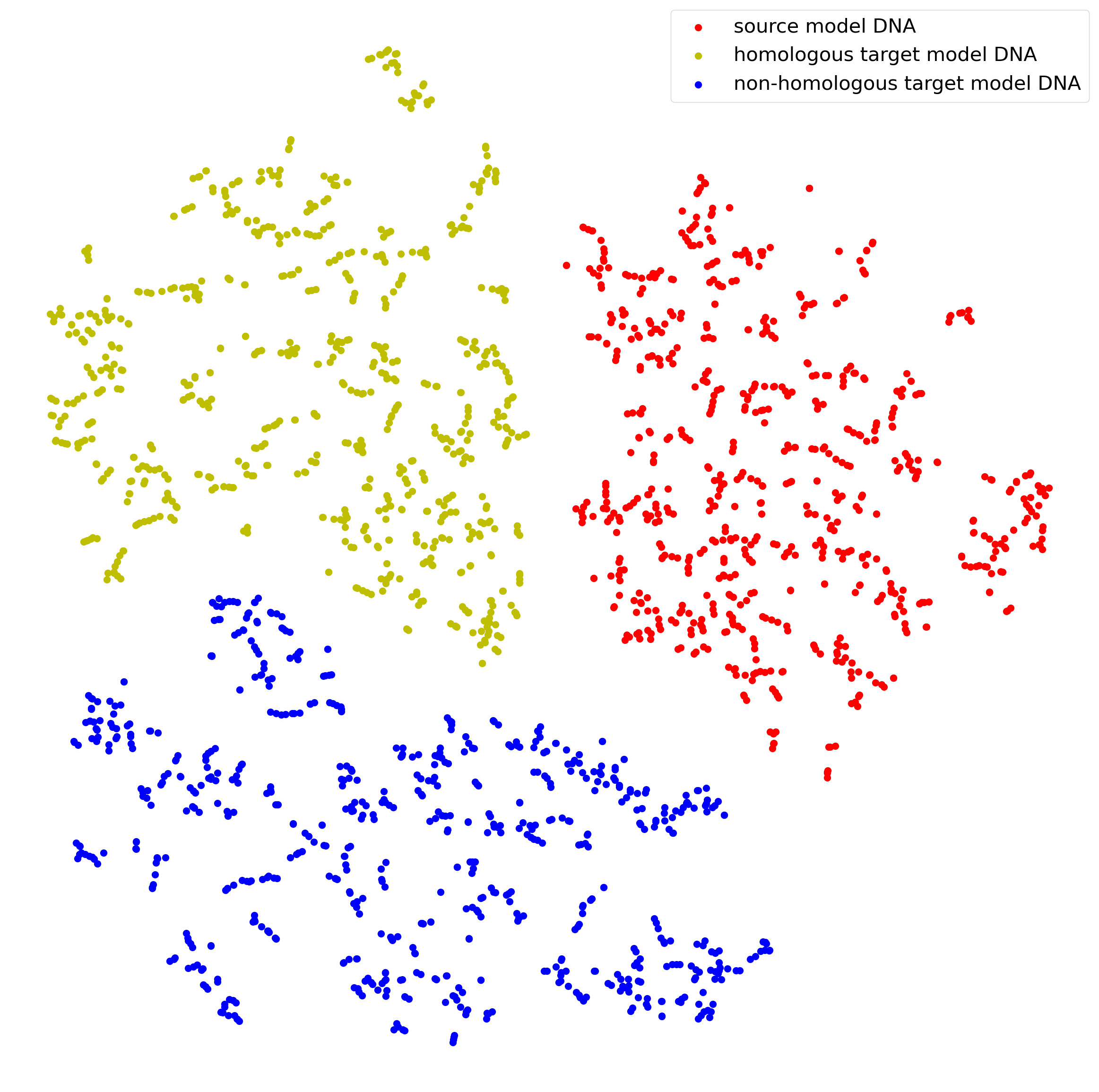}} \ \ 
		\subfigure[]{\includegraphics[width=2cm, height=2cm]{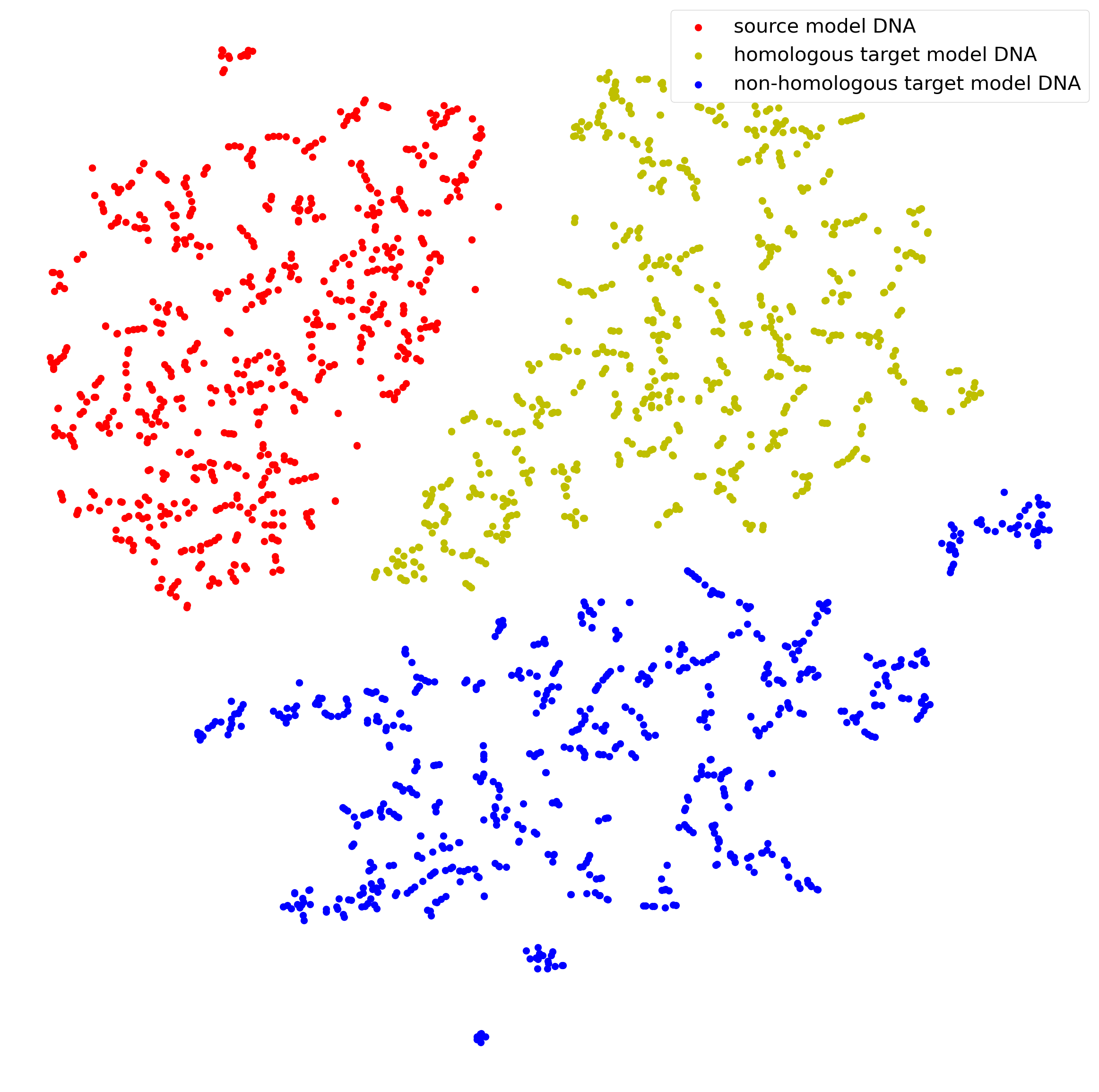}} \ \ 
		\subfigure[]{\includegraphics[width=2cm, height=2cm]{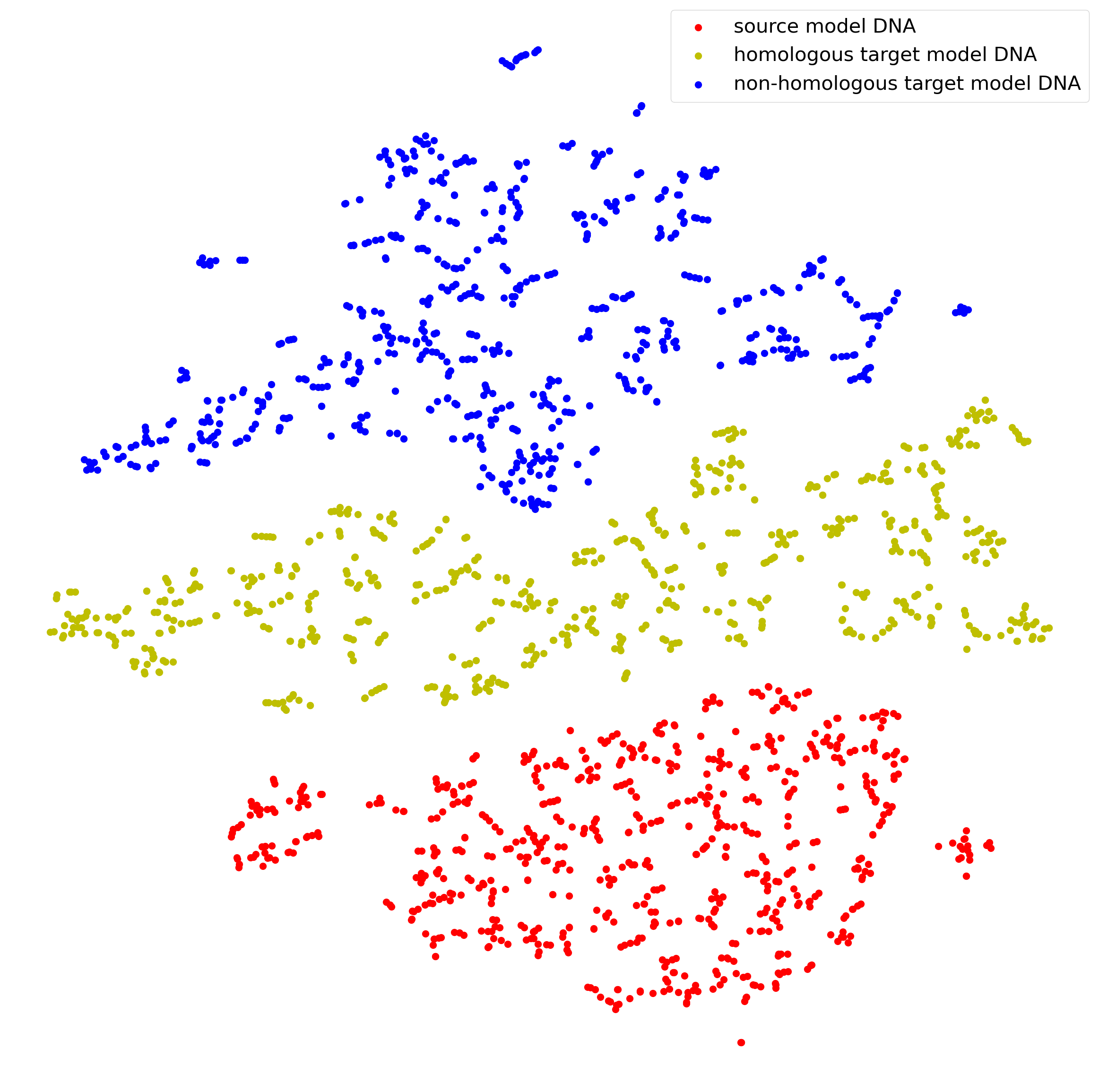}} \\
		\subfigure[]{\includegraphics[width=2cm, height=2cm]{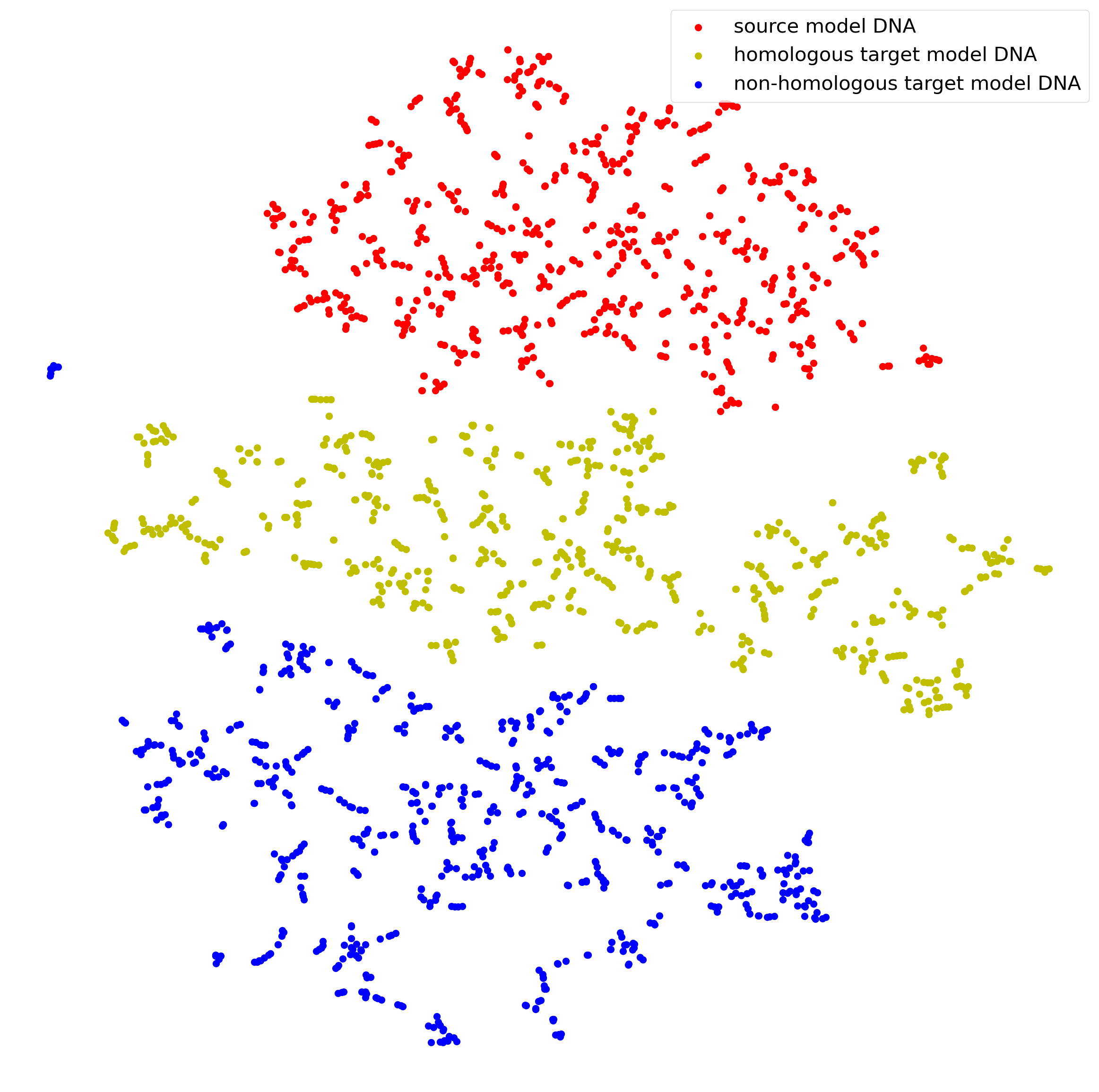}} \ \ 
		\subfigure[]{\includegraphics[width=2cm, height=2cm]{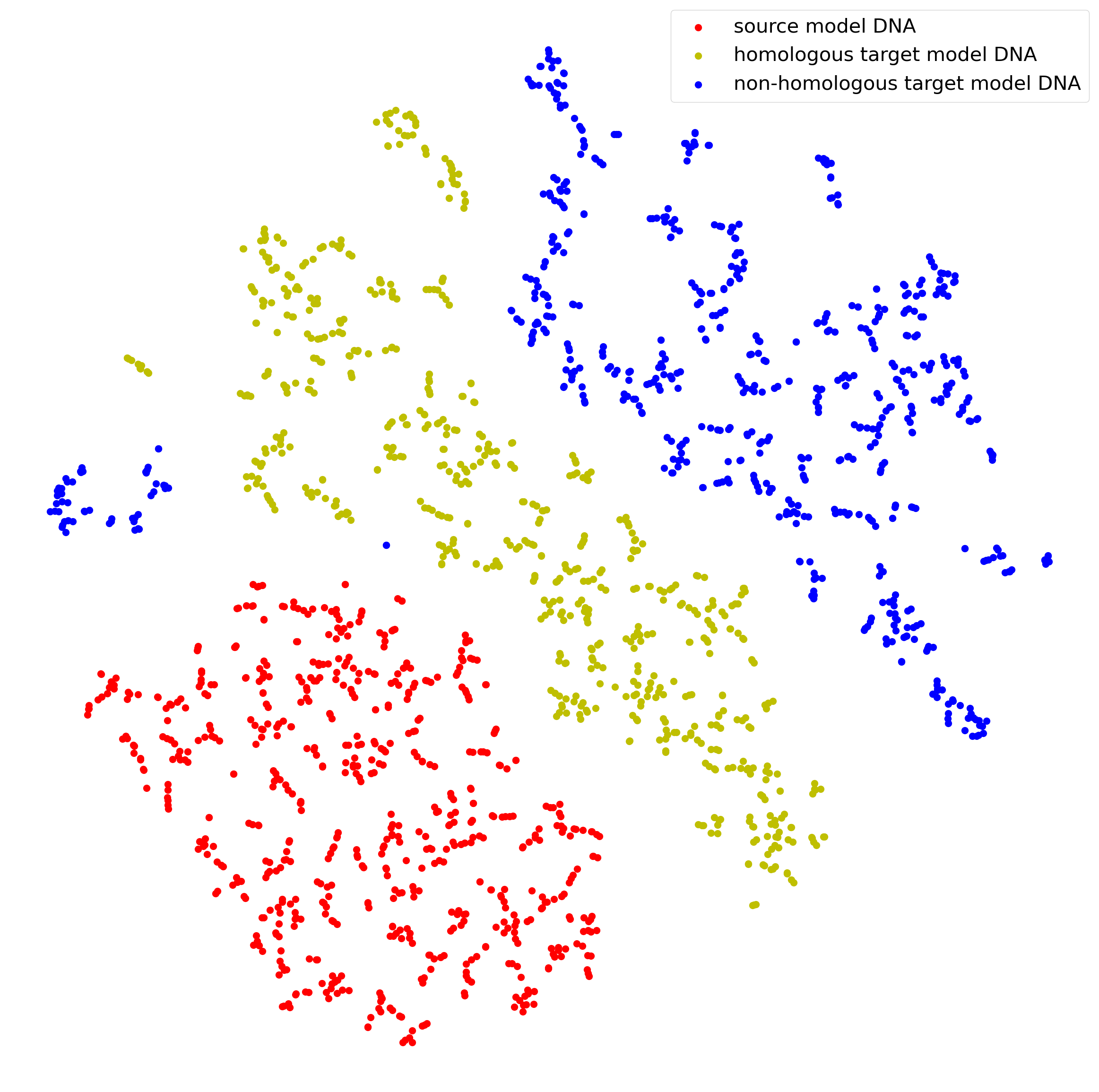}} \ \ 
		\subfigure[]{\includegraphics[width=2cm, height=2cm]{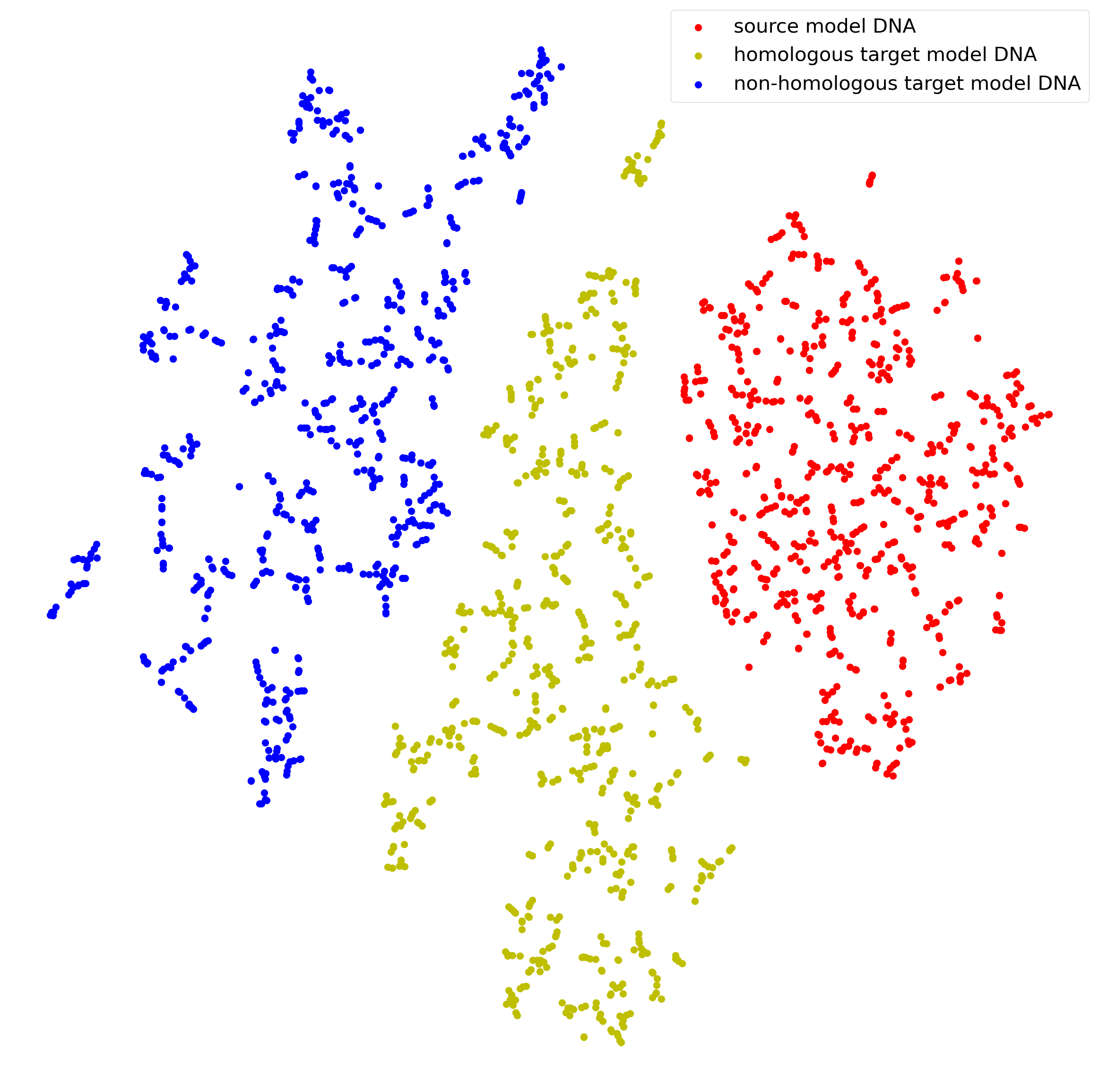}}  
		\subfigure[]{\includegraphics[width=2cm, height=2cm]{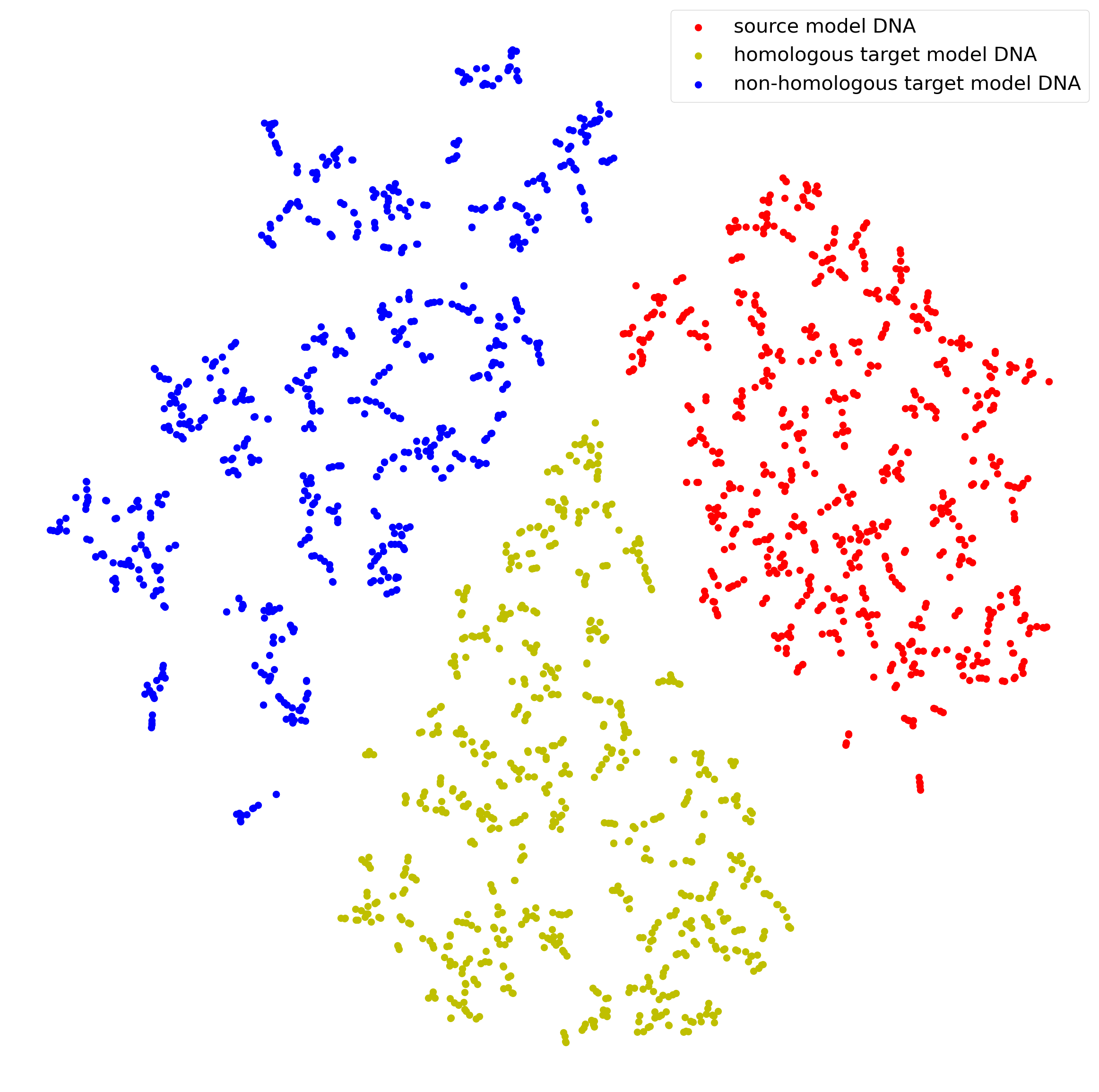}} \ \ 
		\subfigure[]{\includegraphics[width=2cm, height=2cm]{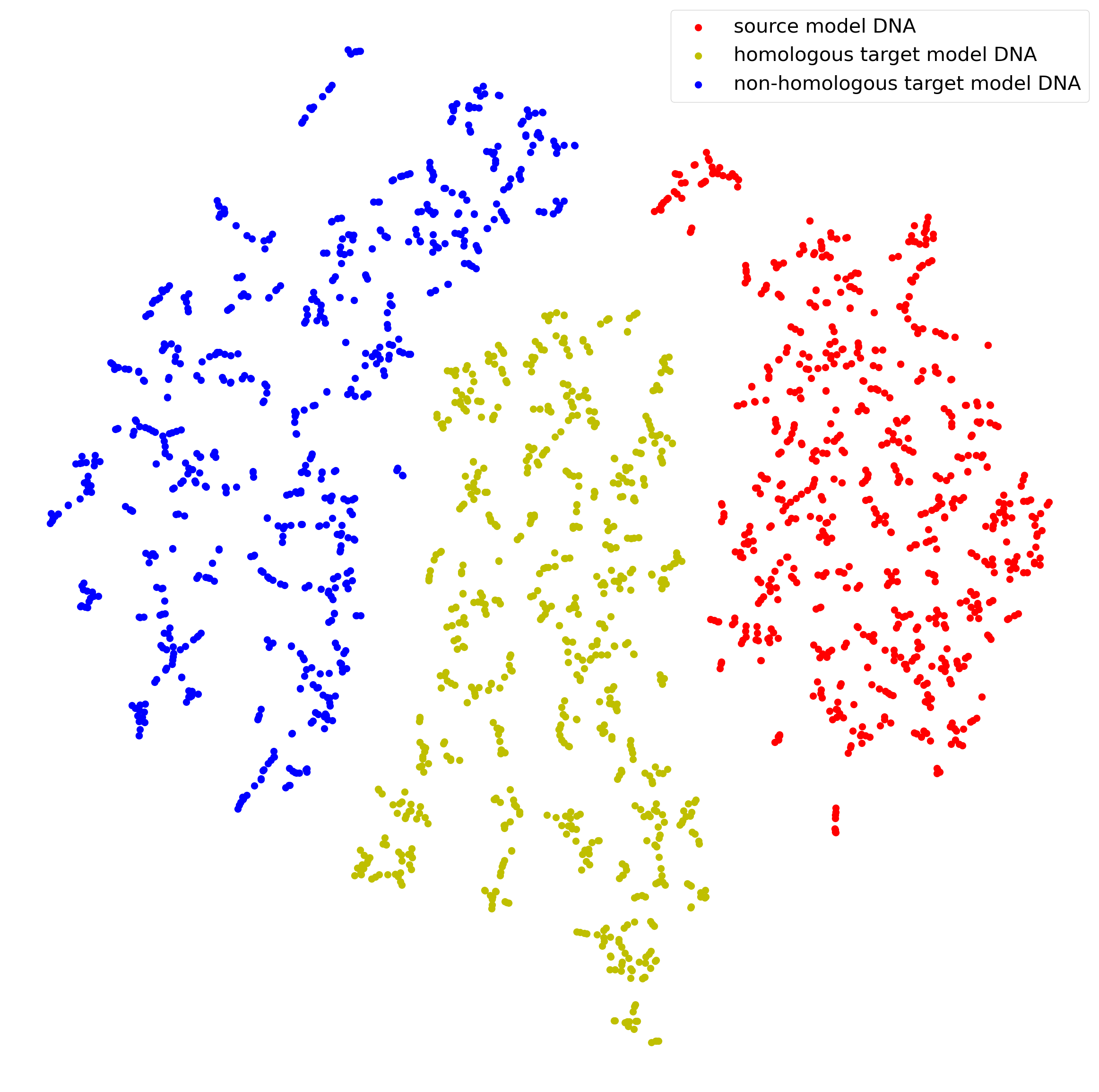}}\\
		\subfigure[]{\includegraphics[width=2cm, height=2cm]{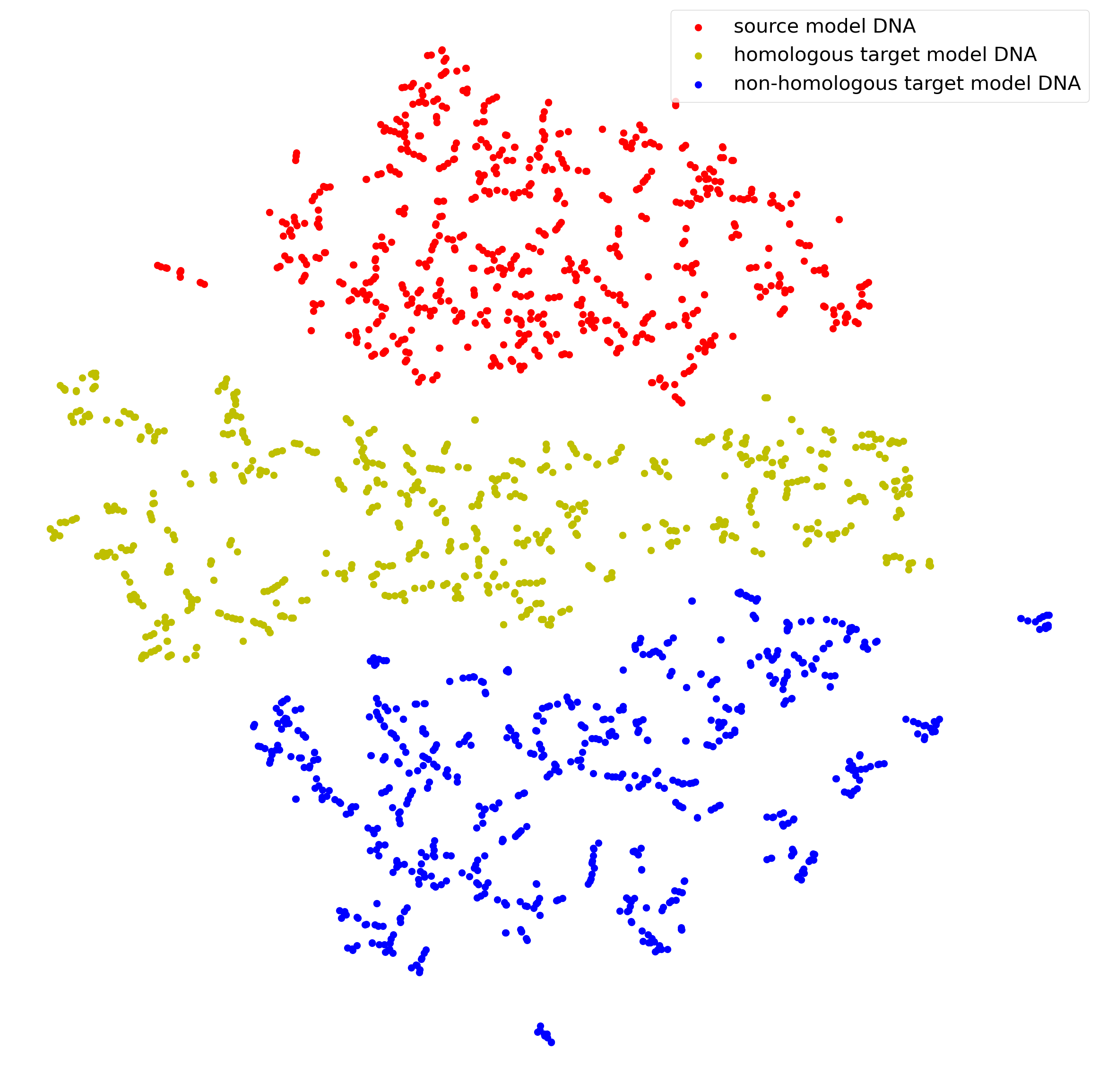}} \ \ 
		\subfigure[]{\includegraphics[width=2cm, height=2cm]{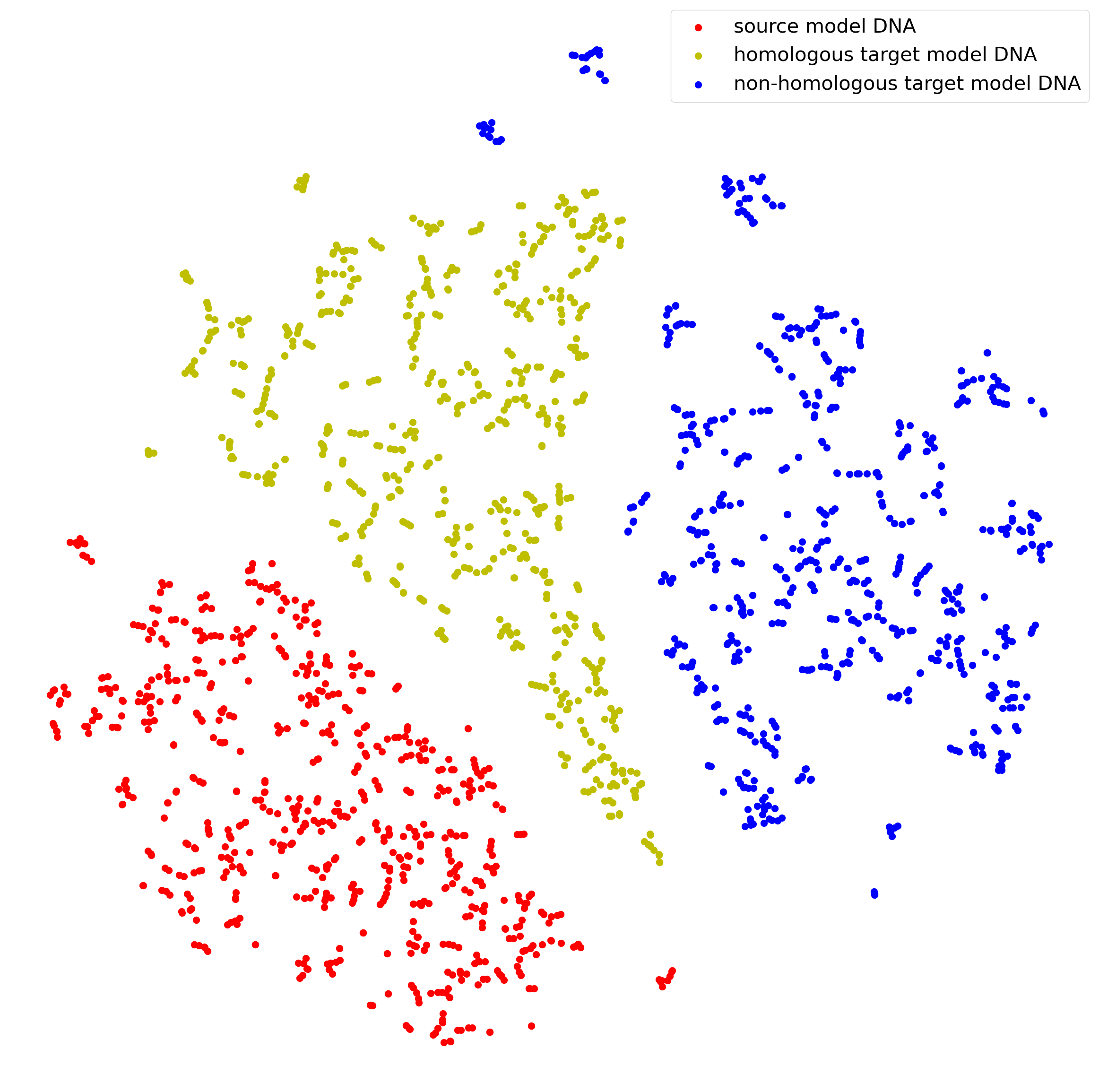}} \ \  \subfigure[]{\includegraphics[width=2cm, height=2cm]{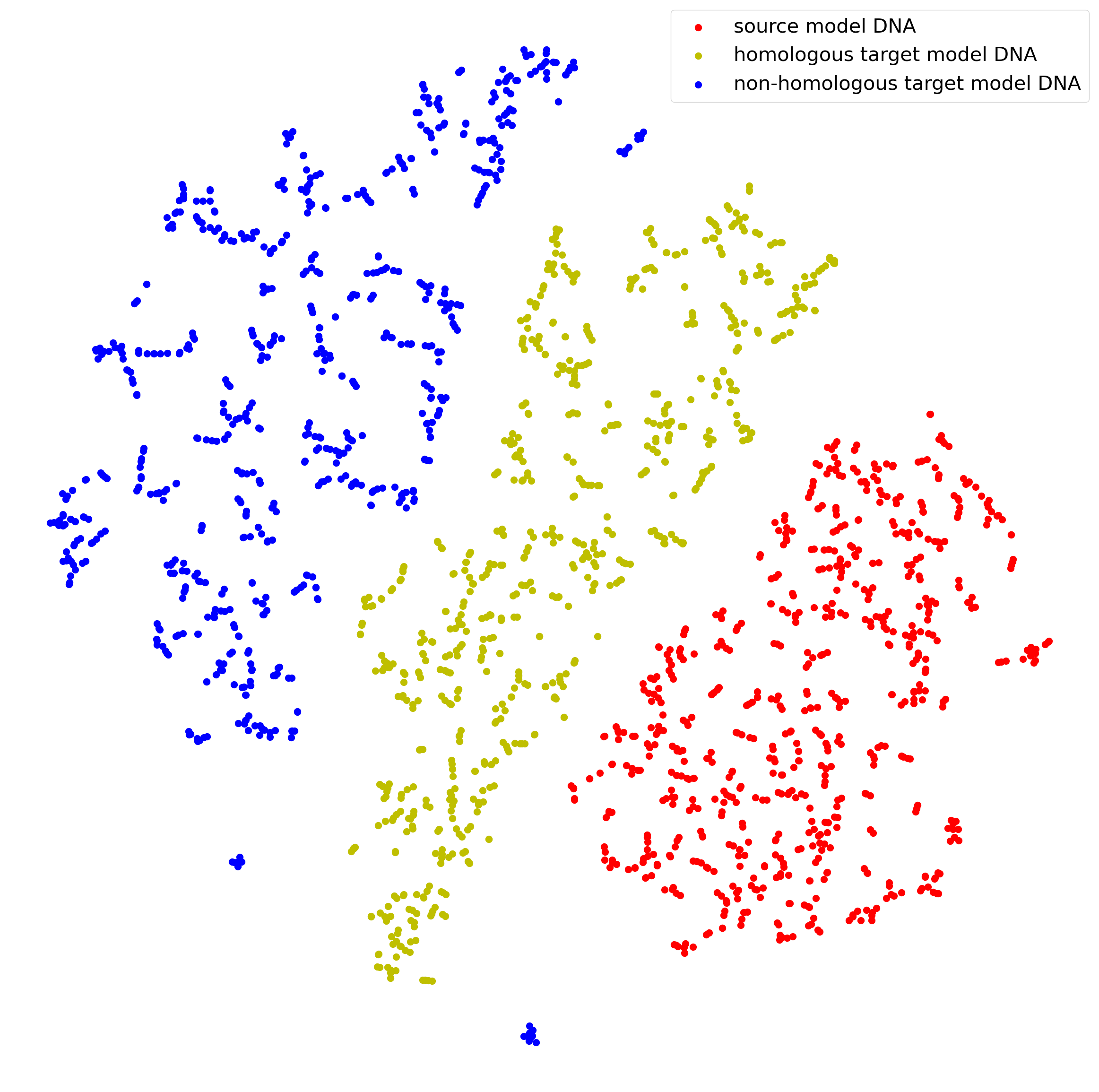}}  \ \ \subfigure[]{\includegraphics[width=2cm, height=2cm]{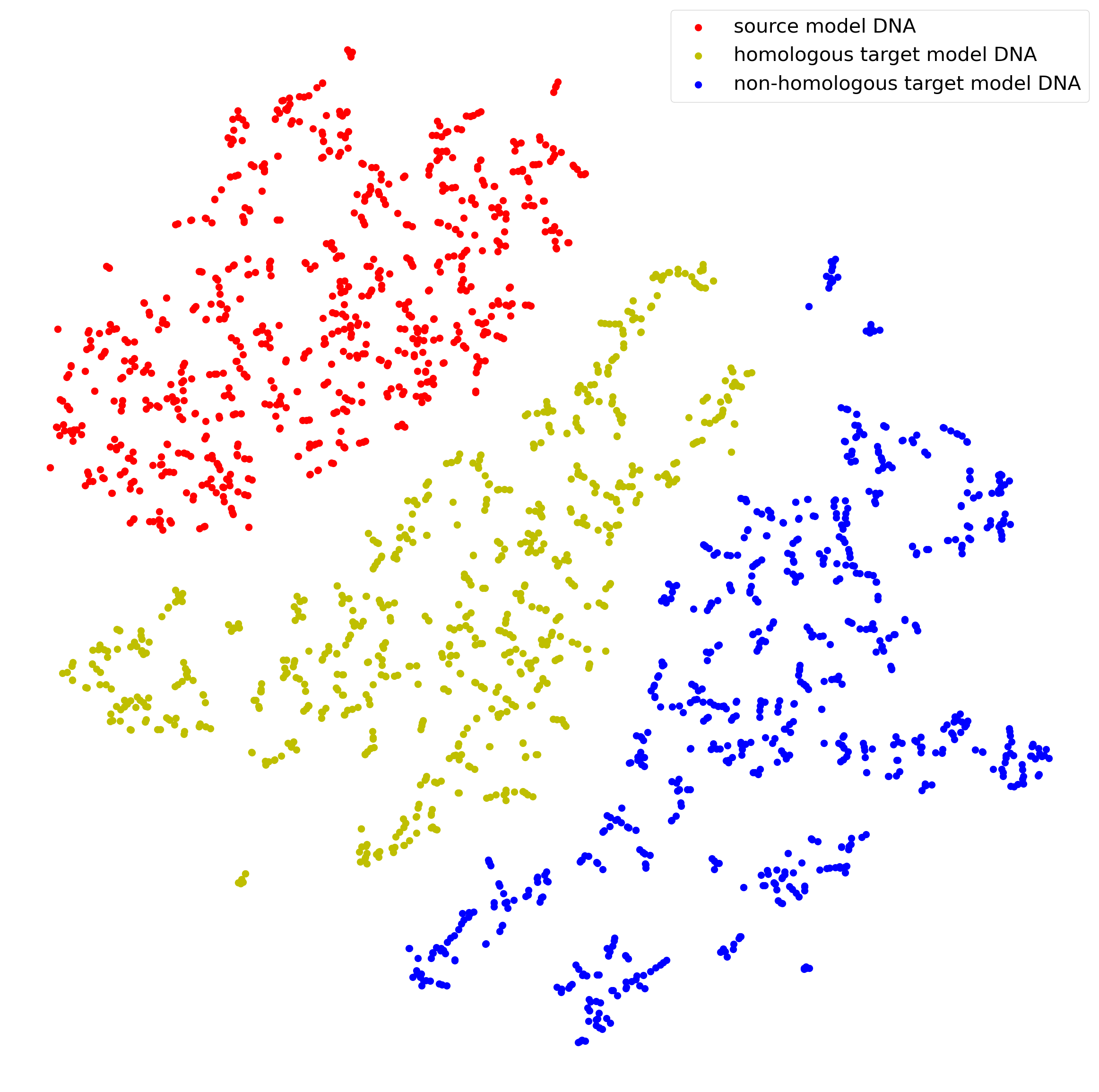}} \ \  \subfigure[]{\includegraphics[width=2cm, height=2cm]{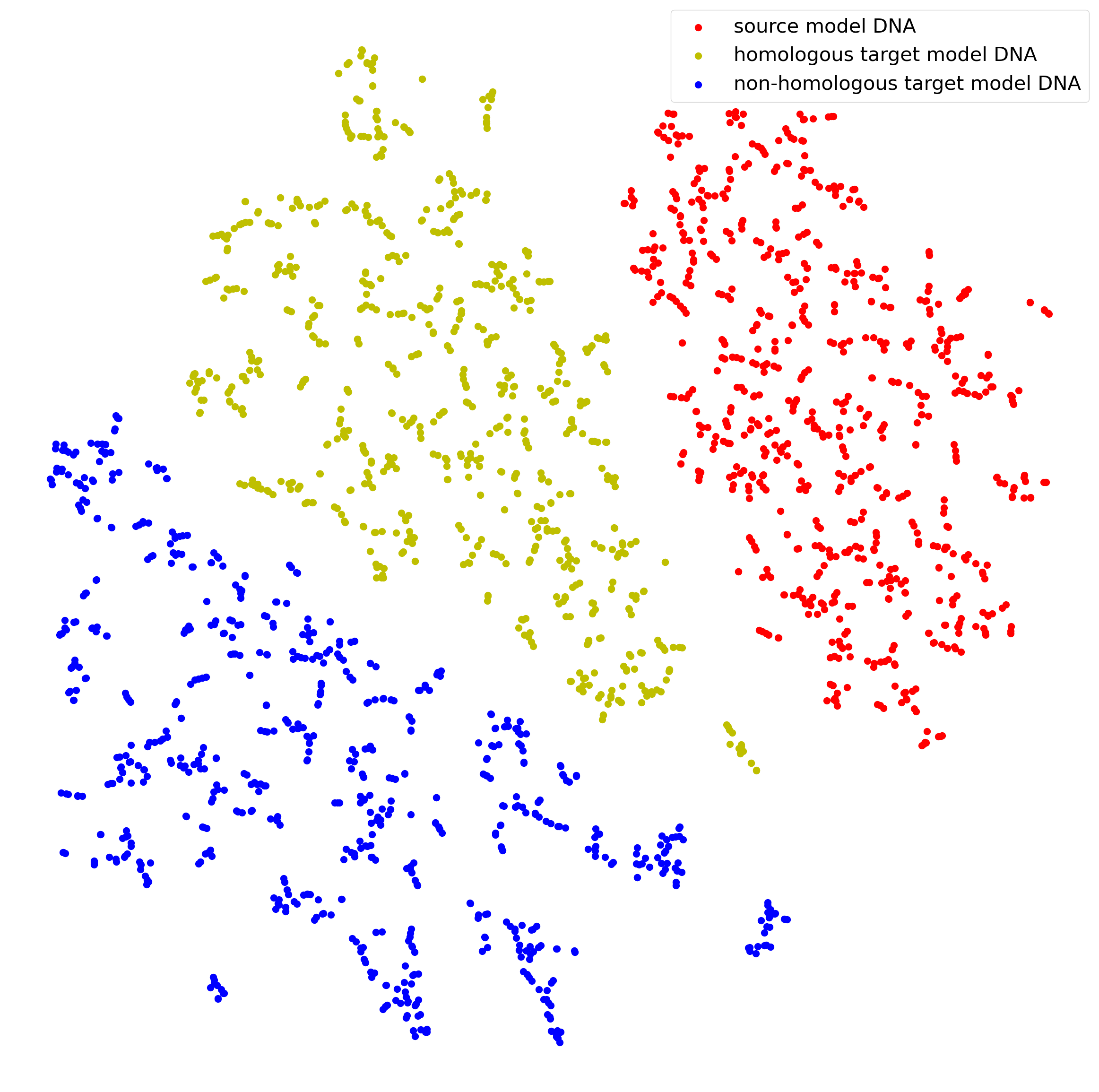}}\\
		\subfigure[]{\includegraphics[width=2cm, height=2cm]{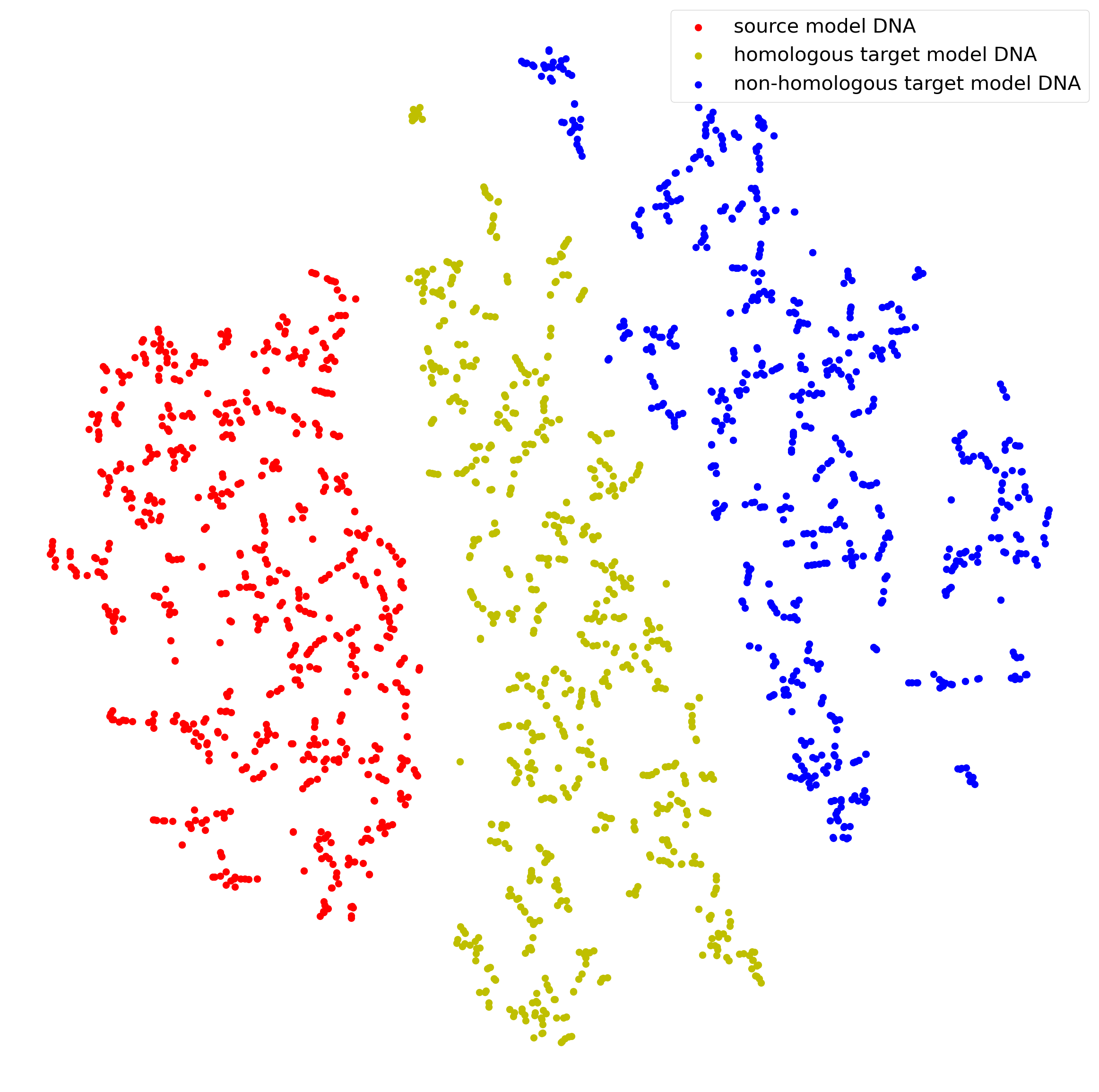}}\ \
		\subfigure[]{\includegraphics[width=2cm, height=2cm]{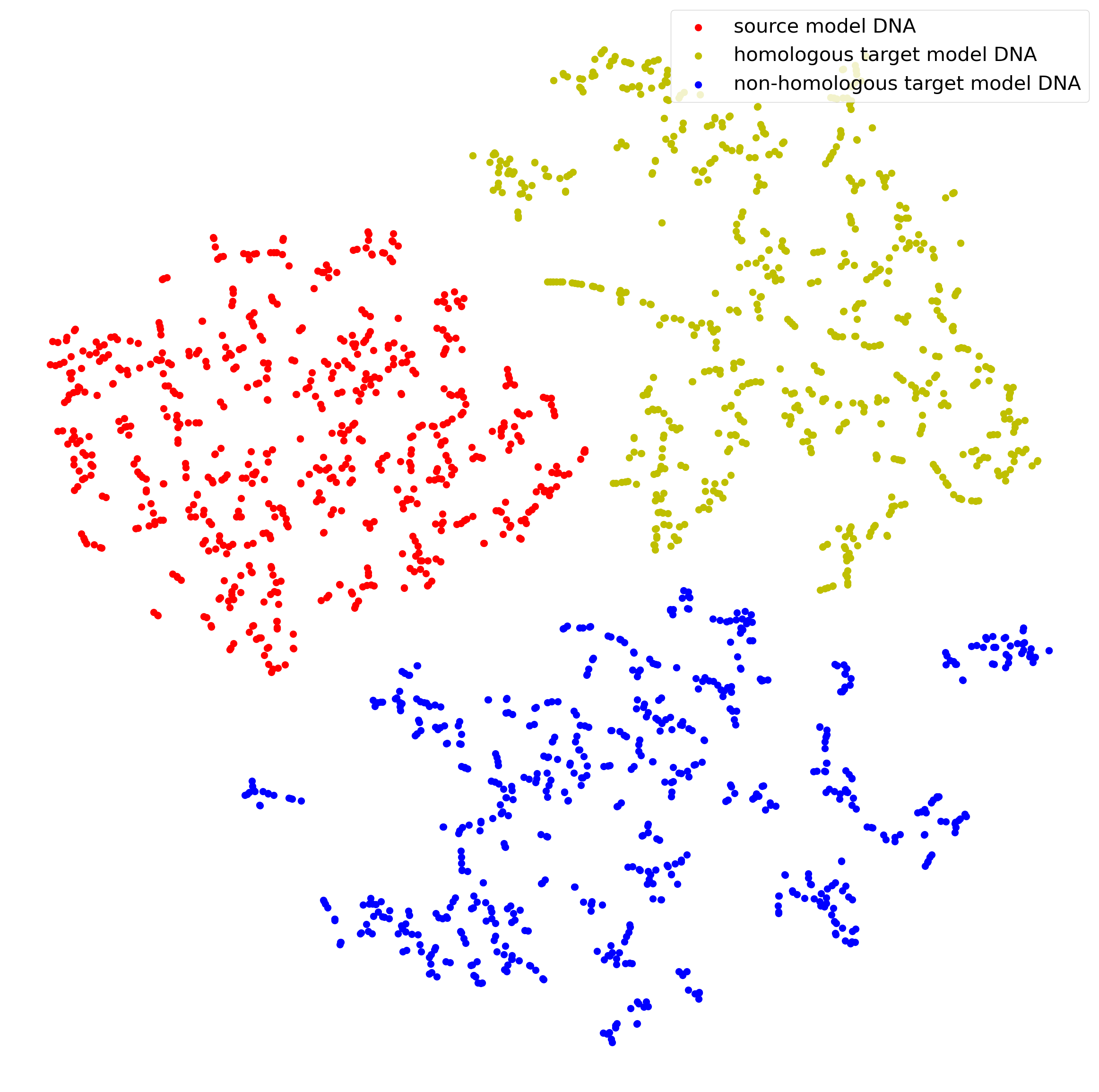}}\ \
		\subfigure[]{\includegraphics[width=2cm, height=2cm]{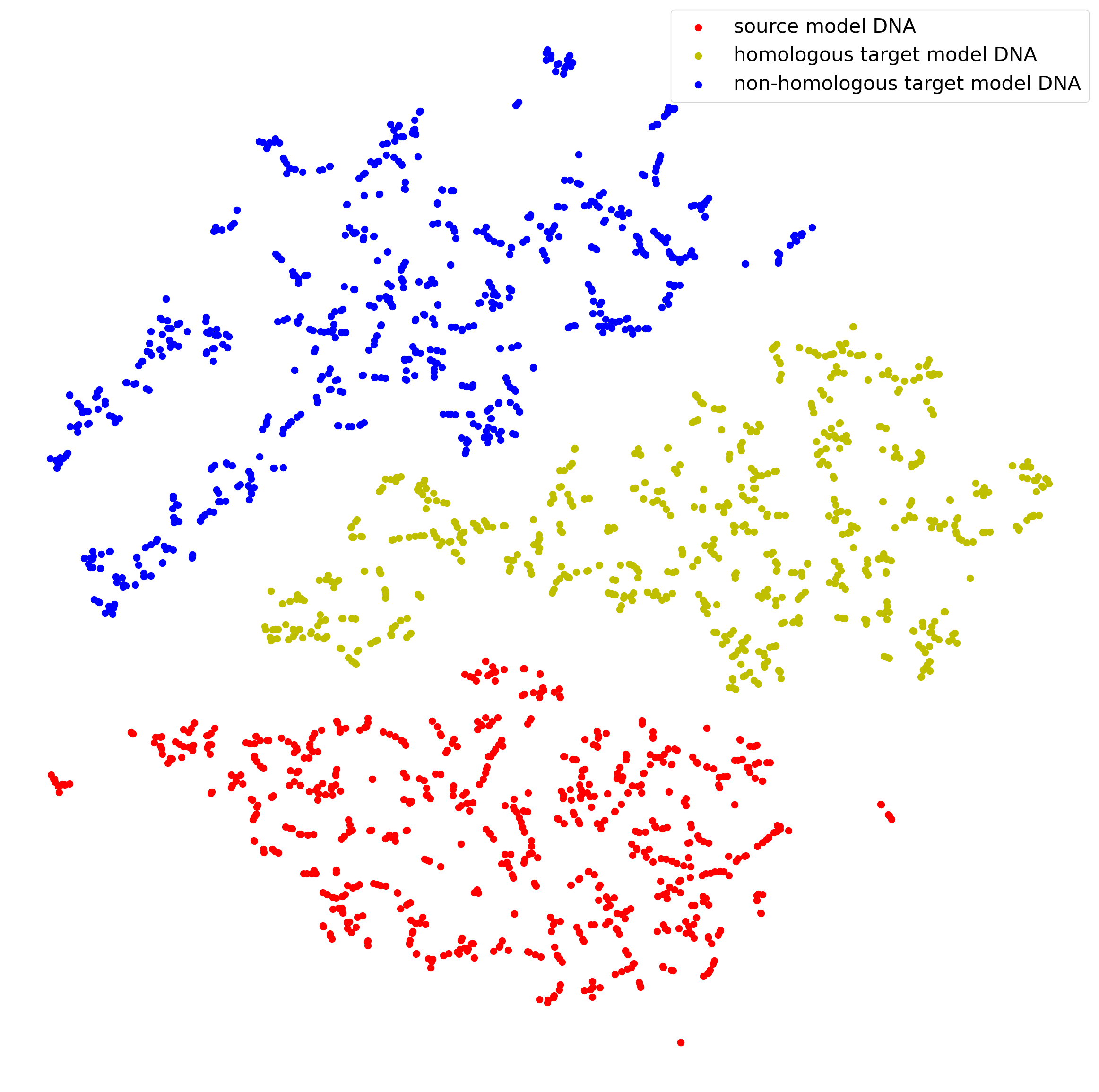}}\ \
		\subfigure[]{\includegraphics[width=2cm, height=2cm]{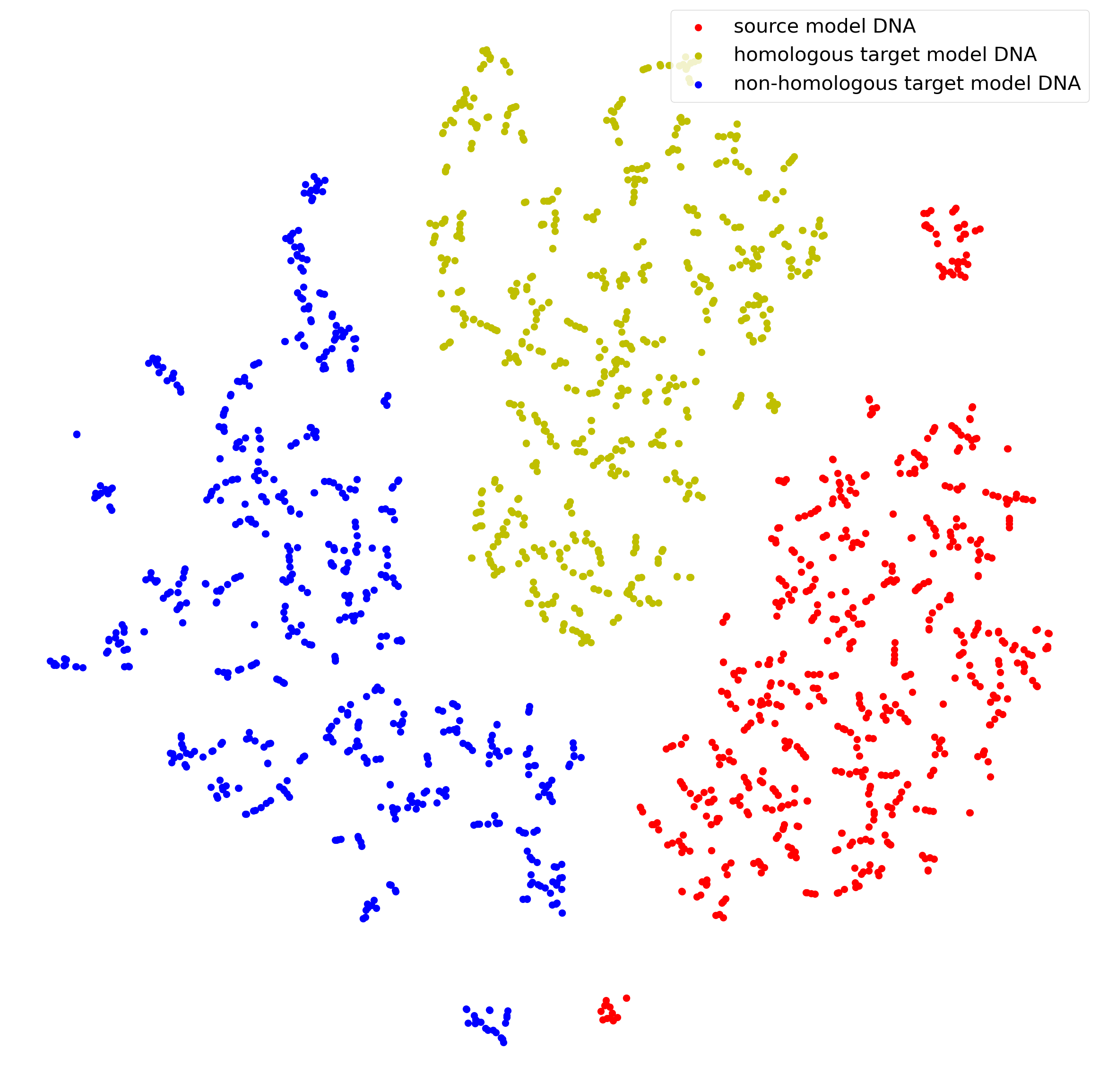}}\ \
		\subfigure[]{\includegraphics[width=2cm, height=2cm]{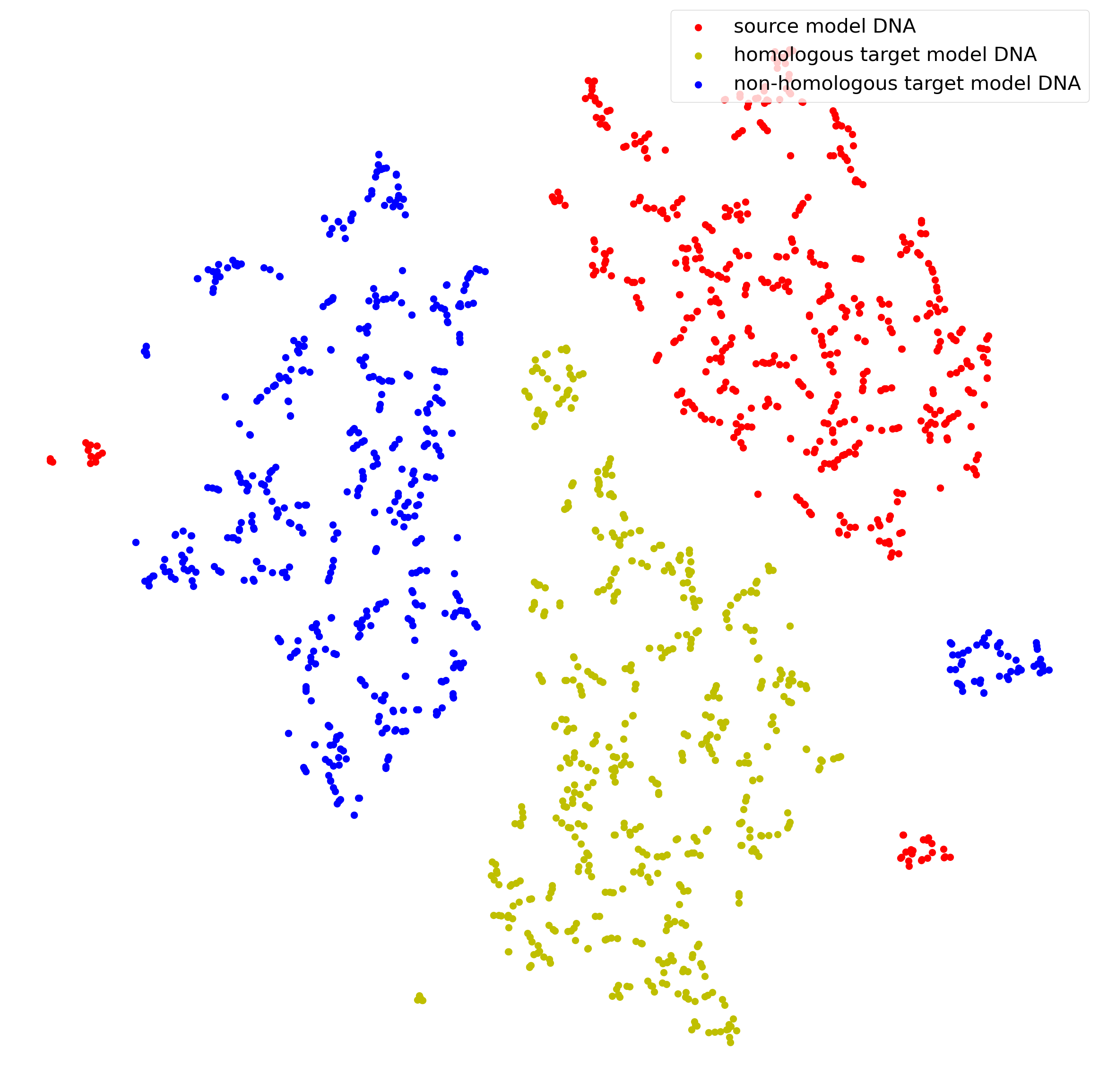}}
		\caption{The visualization of the DNA fragments of the source model (red), homologous target model (yellow), and non-homologous target model (blue).}
		\label{fig:vifornlp}
	\end{figure*}
	
	In addition to the previous visualization on CV task, we also provide visualization on NLP task in Figure \ref{fig:vifornlp}. We visualize the generated DNA fragments of the third NLP experiment (i.e., the third row of Table \ref{sample-tablefull11}).

\end{document}